\documentclass[10pt,twocolumn,letterpaper]{article}

\usepackage{cvpr}
\usepackage{times}
\usepackage{epsfig}
\usepackage{graphicx}
\usepackage{amsmath}
\usepackage{amssymb}

\usepackage[usenames, dvipsnames]{color}
\usepackage[ruled]{algorithm2e}
\usepackage[noend]{algpseudocode}
\usepackage{wrapfig}
\usepackage{tablefootnote}
\usepackage{paralist}  
\usepackage{subfig}
\usepackage[export]{adjustbox}
\usepackage{array}
\usepackage{pbox}
\usepackage{multirow}
\usepackage{booktabs}

\usepackage{soul}

\usepackage{lipsum,multicol}

\newcolumntype{L}[1]{>{\raggedright\let\newline\\\arraybackslash\hspace{0pt}}m{#1}}
\newcolumntype{C}[1]{>{\centering\let\newline\\\arraybackslash\hspace{0pt}}m{#1}}
\newcolumntype{R}[1]{>{\raggedleft\let\newline\\\arraybackslash\hspace{0pt}}m{#1}}

\usepackage[colorlinks=true,bookmarks=false,pagebackref]{hyperref}

\cvprfinalcopy 

\def\cvprPaperID{1023} 

\begin{document}

\title{Automatic adaptation of object detectors to new domains using self-training 
}

\author{Aruni RoyChowdhury \qquad Prithvijit Chakrabarty \qquad Ashish Singh \qquad SouYoung Jin  \\ 
Huaizu Jiang \qquad Liangliang Cao \qquad Erik Learned-Miller\\
College of Information and Computer Sciences\\
University of Massachusetts Amherst\\
{\tt\small \{arunirc, pchakrabarty, ashishsingh, souyoungjin, hzjiang, llcao, elm\}@cs.umass.edu}
}

\maketitle

\begin{abstract}
  This work addresses the unsupervised adaptation of an existing object detector to a new target domain. We assume that a large number of unlabeled videos from this domain are readily available. We automatically obtain labels on the target data by using high-confidence detections from the existing detector, augmented with hard (misclassified) examples acquired by exploiting temporal cues using a tracker. These automatically-obtained labels are then used for re-training the original model.
  A modified knowledge distillation loss is proposed, and we investigate several ways of assigning soft-labels to the training examples from the target domain. 
  Our approach is empirically evaluated on challenging face and pedestrian detection tasks: a face detector trained on WIDER-Face, which consists of high-quality images crawled from the web, is adapted to a large-scale surveillance  data set; a pedestrian detector trained on clear, daytime images from the BDD-100K driving data set is adapted to all other scenarios such as rainy, foggy, night-time. Our results demonstrate the usefulness of incorporating hard examples obtained from tracking, the advantage of using soft-labels via distillation loss versus hard-labels, and show promising performance as a simple method for unsupervised domain adaptation of object detectors, with minimal dependence on hyper-parameters.  Code and models are available at \url{http://vis-www.cs.umass.edu/unsupVideo/}
\end{abstract}

\section{Introduction}
\label{sec:intro}

\begin{figure}[tb]
\centering
\includegraphics[trim=0.8in 0 0.5in 0.1in,clip=true, width=0.5\textwidth]{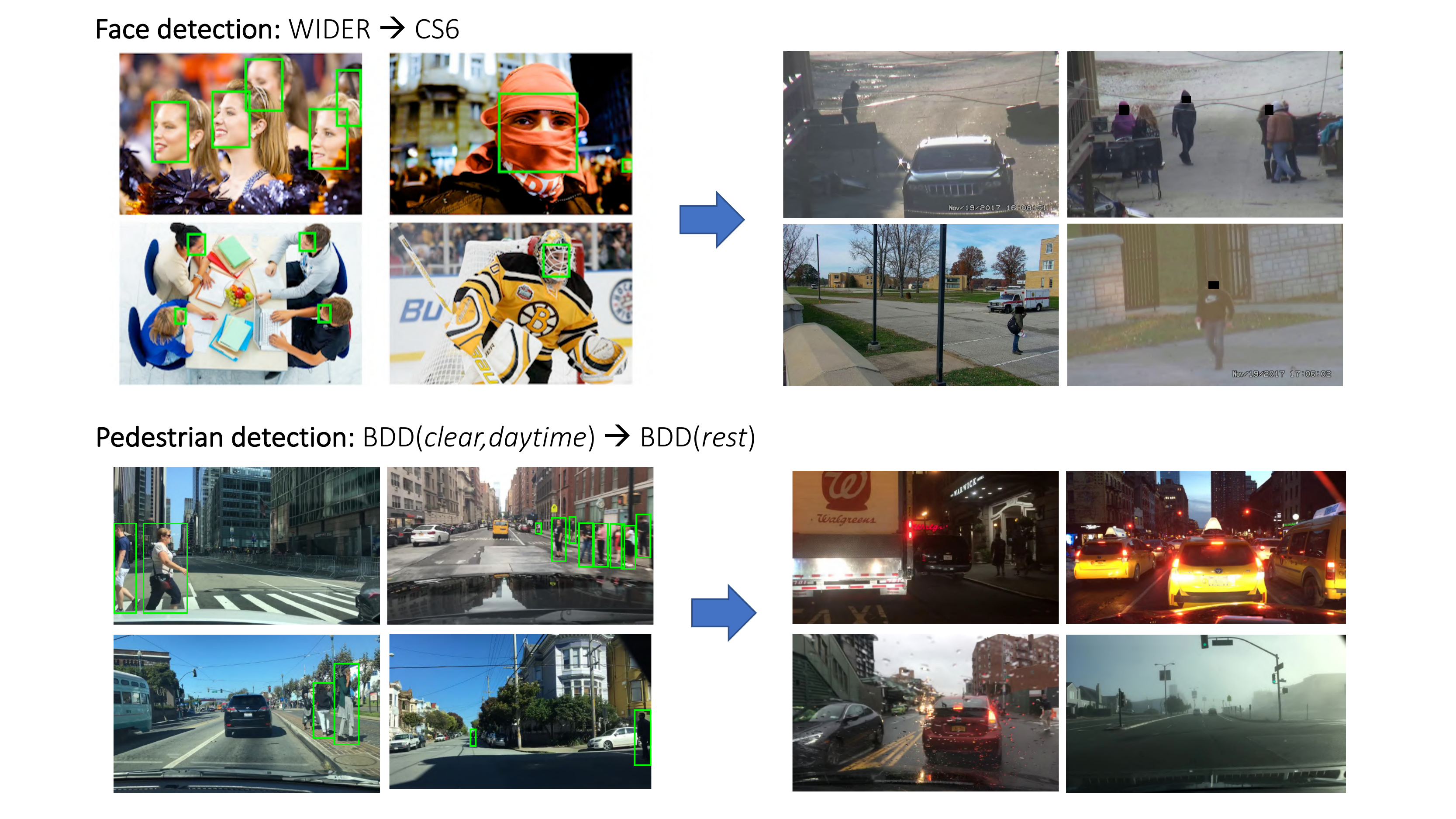} 
\vspace{-8mm}
\caption{\textbf{Unsupervised cross-domain object detection.} \textit{Top:} adapting a face detector trained on labeled high-quality web images from WIDER-Face~\cite{yang16wider} to unlabeled CS6/IJB-S~\cite{kalkaijb}  video frames. \textit{Bottom:} adapting a pedestrian detector trained on labeled images from the \textit{(clear, daytime)} split of the BDD-100k dataset~\cite{yu2018bdd100k} to unlabeled videos from all the other conditions (\eg night-time, foggy, rainy, \etc). }
\label{fig:task_example}
\vspace{-4mm}
\end{figure}



The success of deep neural networks has resulted in state-of-the-art object detectors that obtain high accuracy on standard vision benchmarks (\eg MS-COCO~\cite{lin2014microsoft}, PASCAL VOC~\cite{Everingham10}, etc.), and are readily available for download as out-of-the-box detection models~\cite{Detectron2018,huang2017speed}. However, it is unrealistic to expect a single detector to generalize to \textit{every} domain. Due to the data-hungry nature of supervised training of deep networks, it would require a lot of labeling efforts to re-train a detector in a completely supervised manner for a new scenario.

This paper considers the following problem: Given an off-the-shelf detector, can we let it automatically improve itself by watching a video camera? 
We hope to find a new algorithm based on unsupervised self-training that leverages large amounts of readily-available unlabeled video data, so that it can relieve the requirement of labeling effort for the new domain, which is tedious, expensive, and difficult to scale up. Such a solution may be very useful to generalize existing models to new domains without supervision, \eg a pedestrian detection system trained on imagery of US streets can adapt to cities in Europe or Asia, or help an off-the-shelf face detector improve its performance on video footage. Such an algorithm would be a label efficient solution for large-scale domain adaptation, obviating the need for costly bounding-box annotations when faced with a new domain.


Recent approaches to unsupervised domain adaptation in deep networks have attempted to learn domain invariant features through an adversarial domain discriminator~\cite{chen2018domain,ganin2014unsupervised,ganin2016domain,tzeng2017adversarial,hoffman2017cycada}, or by transforming labeled source images to resemble the target domain using a generative adversarial network (GAN)~\cite{inoue2018cross,zhu2017unpaired,bousmalis2017unsupervised}. 
\textit{Self-training} is a comparatively simpler alternate strategy, where the off-the-shelf model's predictions on the new domain are regarded as ``pseudo-labeled'' training samples~\cite{lee2013pseudo,chapelle2009semi,blum1998combining,westonlarge}; however this approach would involve re-training using significantly noisy labels. It becomes even more challenging when we consider object detectors in particular, as the model may consider a wrongly-labeled instance as a hard example~\cite{shrivastava2016training} during training, and expend a lot of efforts trying to learn it.

In this paper, we leverage two types of information that is useful for object detection. First, 
object detectors can benefit from learning the temporal consistency in videos. 
Some hard cases missed by the detector could be recognized if the object is detected in neighboring frames. We combine both tracking and detection into one framework, and automatically refine the labels based on detection and tracking results.  
Second, there are examples of varying difficulty in the new domain, and we propose a distillation-based loss function to accommodate this relative ordering in a flexible fashion. We design several schemes to assign soft-labels to the target domain samples, with minimal dependence on hyper-parameters.
We evaluate our methods for improving \textit{single-image} detection performance without labels on challenging face and pedestrian detection tasks, where the target domain contains a large number of unlabeled videos. Our results show that training with soft labels improves over the usual hard (\ie 0 or 1) labels, and reaches comparable to better performance relative to adversarial methods without extra parameters.
The paper is organized as follows -- relevant literature is reviewed in Sec.~\ref{sec:related}, the proposed approach is described in Sec~\ref{sec:method} and experimental results are presented in Sec~\ref{sec:experiments}.

\section{Related Work}
\label{sec:related}

\noindent
\textbf{Semi-supervised learning.} Label-efficient semi-supervised methods of training object recognition models have a long history in computer vision~\cite{rosenberg2005semi,weber2000unsupervised,baluja1999probabilistic,selinger2001minimally,levin2003unsupervised,fergus2003object}. For a survey and empirical comparison of various semi-supervised learning methods applied to deep learning, we refer the reader to Odena~\etal~\cite{odena2018realistic}. We focus on the \textit{self-training} approach~\cite{chapelle2009semi,blum1998combining,westonlarge,lee2013pseudo}, which involves creating an initial baseline model on fully labeled data and then using this model to estimate labels on a novel weakly-labeled or unlabeled dataset. A subset of these estimated labels that are most likely to be correct are selected and used to re-train the baseline model, and the process continues in an incremental fashion~\cite{nigam2000analyzing,li2005setred,muhlenbach2004identifying,jiang2004editing,wilson1972asymptotic}. 
 In the context of object detection, Rosenberg~\etal~\cite{rosenberg2005semi} used the detections from a pre-trained object detector on unlabeled data as pseudo-labels and then trained on a subset of this noisy labeled data in an incremental re-training procedure. 
 Recently, the \textit{data distillation} approach~\cite{radosavovic2017data} aimed to improve the performance of fully-supervised state-of-the-art detectors by augmenting the training set with massive amounts of pseudo-labeled data. In their case, the unlabeled data was from the same domain as the labeled data, and pseudo-labeling was done by selecting the predictions from the baseline model using test-time data augmentation. Jin~\etal~\cite{jin2018unsup} use tracking in videos to gather \textit{hard examples} -- \ie objects that fail to be detected by an object detector (false negatives); they re-train using this extra data to improve detection on still images. Our work shares the latter's strategy of exploiting temporal relationships to automatically obtain hard examples, but our goal is fundamentally different -- we seek to \textit{adapt} to a new target domain, while Jin~\etal use the target domain to mine extra training samples to improve performance back in the source domain. We note that improvements in network architecture specific to video object recognition~\cite{feichtenhofer2017detect,wang2018fully} are orthogonal to our current motivation. 
 
 \noindent \textbf{Hard examples.} Emphasizing difficult training samples has been shown to be useful in several works -- \eg online hard example mining (OHEM)~\cite{shrivastava2016training}, boosting~\cite{schapire1998boosting}. Weinshall and Amir~\cite{weinshall2018theory} show that for certain problem classes, when we do not have access to an optimal hypothesis (\eg a teacher), training on examples the current model finds difficult is more effective than a self-paced approach which trains first on easier samples.

\noindent
\textbf{Unsupervised domain adaptation.} There has been extensive work in addressing the shift between source and target domains~\cite{gretton2008dataset,ben2010theory,sugiyama2017dataset} (see Csurka~\cite{csurka2017domain} for a recent survey). Some approaches try to minimize the Maximum Mean Discrepancy~\cite{gretton2008dataset,tzeng2014deep,long2015learning} or the CORAL metric~\cite{sun2016deep} between the distribution of features from the two domains. Another popular direction is an adversarial setup, explored by recent works such as ADDA~\cite{tzeng2017adversarial}, CyCADA~\cite{hoffman2017cycada}, gradient reversal layer (ReverseGrad)~\cite{ganin2016domain,ganin2014unsupervised}, wherein the discriminator tries to predict the domain from which a training sample is drawn, and the model attains domain invariance by trying to fool this discriminator, while also learning from labeled source samples. In particular, the work of Tzeng~\etal~\cite{tzeng2015simultaneous} obtains soft-labels from model posteriors on \textit{source domain} images, aiming to transfer inter-category correlations information across domains. Our soft-labels, on the other hand, are obtained on the \textit{target domain}, have only a single category (therefore inter-class information is not applicable), and aims at preserving information on the relative difficulty of training examples across domains.

\noindent
\textbf{Cross-domain object detection.} 
The domain shift~\cite{kalogeiton2016analysing} of detectors trained on still images and applied to video frames has been addressed in several works, mostly relying on some form of weak supervision on the target domain and selecting target samples based on the baseline detector confidence score~\cite{han2012detection,tang2012shifting,sharma2013efficient,donahue2013semi,kuznetsova2015expanding,chandaadapting}. Several approaches have used weakly-labeled video data for re-training object detectors~\cite{kalal2010pn,singh2016track,tang2012shifting}. Our work is motivated in particular by Tang~\etal~\cite{tang2012shifting}, who use tracking information to get pseudo-labels on weakly-labeled video frames and adopt a curriculum-based approach, introducing easy examples (\ie having low loss) from the target video domain into the re-training of the baseline detector. Despite the common motivation, our work differs on two major points -- (a) we show the usefulness of combining \textit{both} hard and easy examples from the target domain when re-training the baseline model, and (b) using the knowledge distillation loss to counter the effect of label noise.
Jamal~\etal~\cite{Jamal2018CVPR} address the domain shift between various face detection datasets by re-calibrating the final classification layer of face detectors using a residual-style layer in a low-shot learning setting. 
Two recent methods~\cite{inoue2018cross,chen2018domain} for \textit{domain-adaptive object detection} are particularly relevant to our problem.
The weakly-supervised method of Inoue~\etal~\cite{inoue2018cross} first transforms the labeled source (natural) images to resemble the target images (watercolors) using the CycleGAN~\cite{zhu2017unpaired}, fine-tunes the baseline (pre-trained) detector on this  ``transformed source'' data, and then obtains pseudo-labels on the target domain using this domain-adapted model. The task of image generation is fairly difficult, and we posit that it may be possible to address domain adaptation without requiring a generative model as an intermediate step. 
The fully unsupervised method of Chen~\etal~\cite{chen2018domain} learns a domain-invariant representation by using an adversarial loss from a \textit{domain discriminator}~\cite{ganin2014unsupervised,ganin2016domain} at various levels of the Faster R-CNN architecture, showing significant improvements when adapting to challenging domain shifts such as clear to foggy city scenes, simulated to real driving videos, \etc. While a powerful approach, the design of new discriminator layers and adversarial training are both challenging in practice, especially without a labeled validation set on the target domain (as is the case in an unsupervised setting). 
\section{Proposed Approach}
\label{sec:method}

\label{sec:method}
\noindent
Automatically labeling the target domain is described in Sec.~\ref{sec:pseudo-label}, re-training using these pseudo-labels in Sec.~\ref{sec:default_loss} and creating soft-labels in Sec.~\ref{sec:distill}.

\subsection{Automatic Labeling of the Target Domain}
\label{sec:pseudo-label}


\noindent
Self-labeling~\cite{triguero2015self} or pseudo-labeling~\cite{lee2013pseudo} adapts a pre-existing or \textit{baseline} model, trained on a labeled \textit{source} domain $\mathcal{S}$, to a novel unlabeled \textit{target} domain $\mathcal{T}$, by treating the model's own predictions on the new dataset as training labels. In our case, we obtain target domain pseudo-labels by selecting high-confidence predictions of the baseline detector, followed by a refinement step using a tracker.

\noindent
\textbf{Pseudo-labels from detections.} 
The baseline detector is run on every frame of the unlabeled videos in the target domain and if the (normalized) detector confidence score for the $i$-th prediction (\ie the model's posterior), $d_i$, is higher than some threshold $\theta$, then this prediction is added to the set of pseudo-labels.
In practice, we select 0.5 for $\theta$ for face detection and 0.8 for person detection. Note that such a threshold is easily selected by visually inspecting a small number of unlabeled videos from $\mathcal{T}$ (5 videos); we compare with a fully-automated procedure in Sec.~\ref{sec:threshold}.

\begin{figure*}[tbp]
\centering
\begin{tabular}{cc}
    \includegraphics[width=0.2\textwidth]{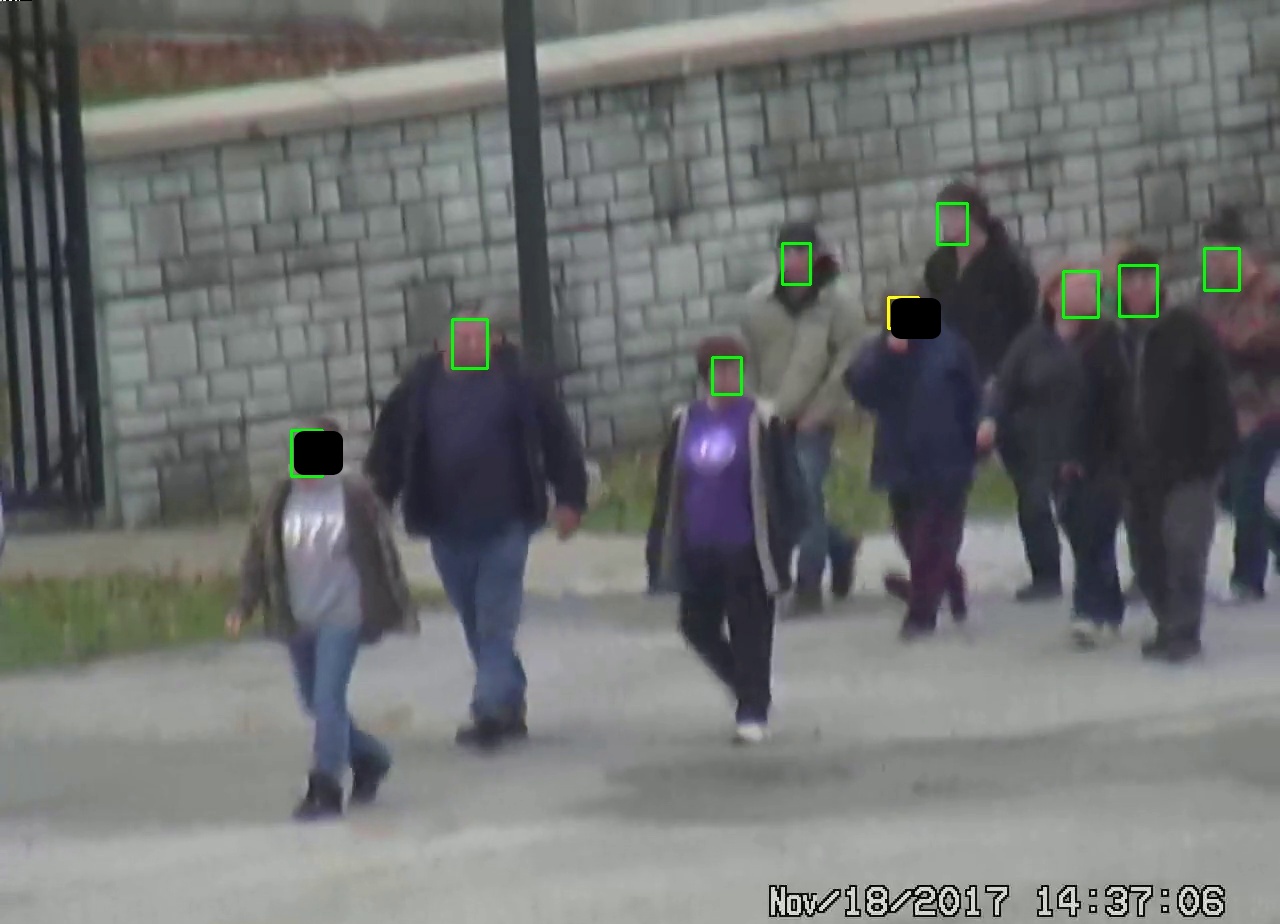} 
    \includegraphics[width=0.2\textwidth]{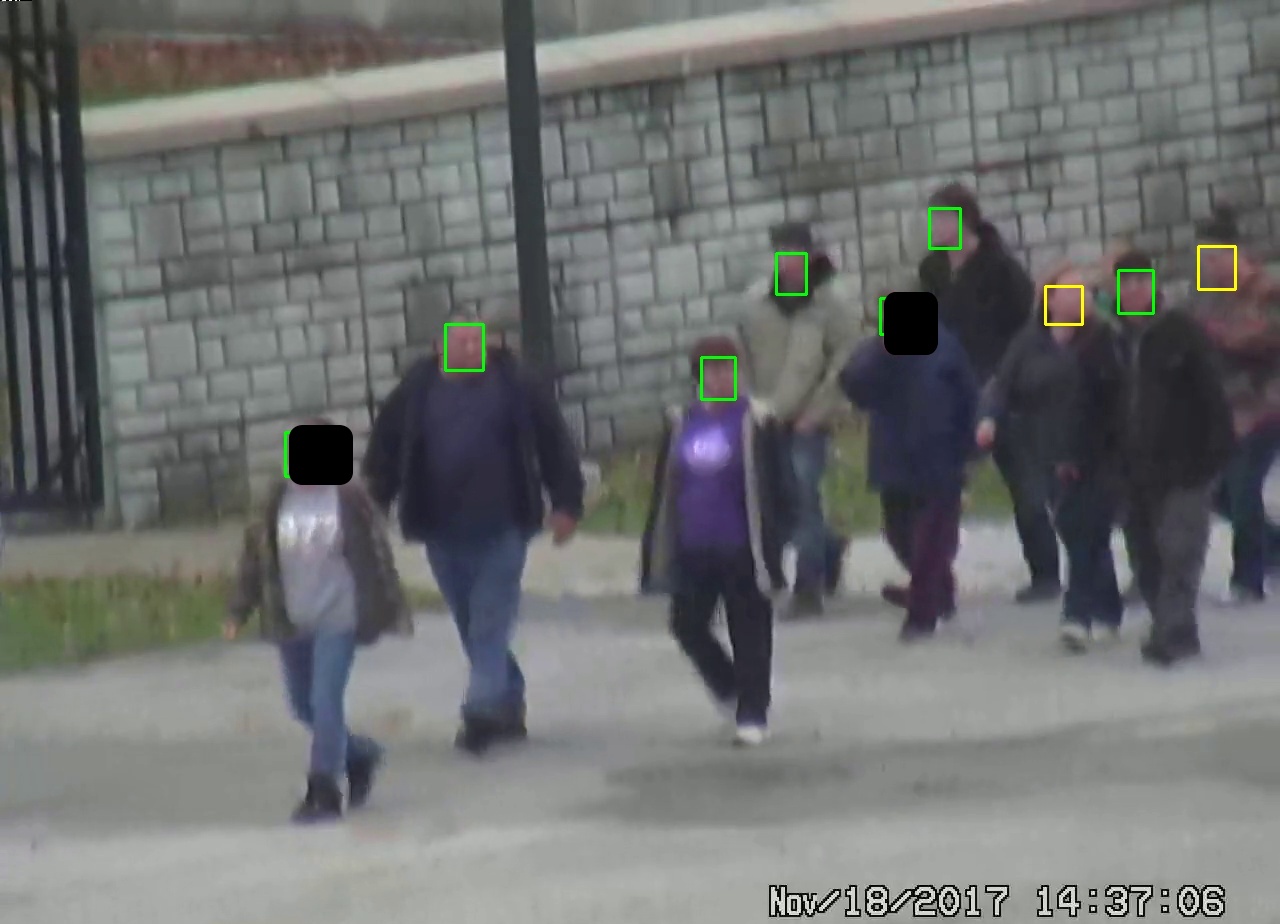} 
    \includegraphics[width=0.2\textwidth]{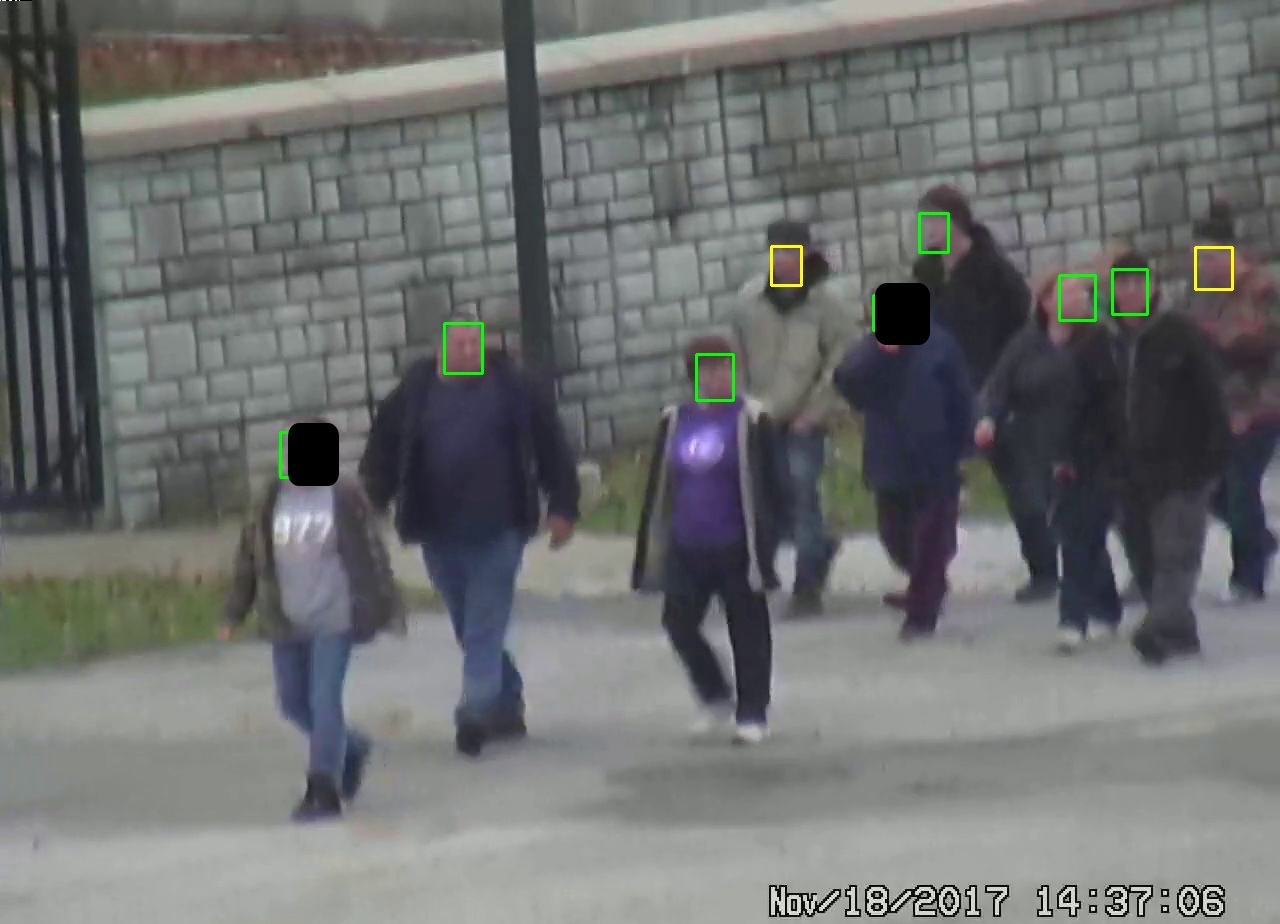}
    &
    \includegraphics[trim=1in 0.5in 1.5in 0,clip=true,width=0.24\textwidth]{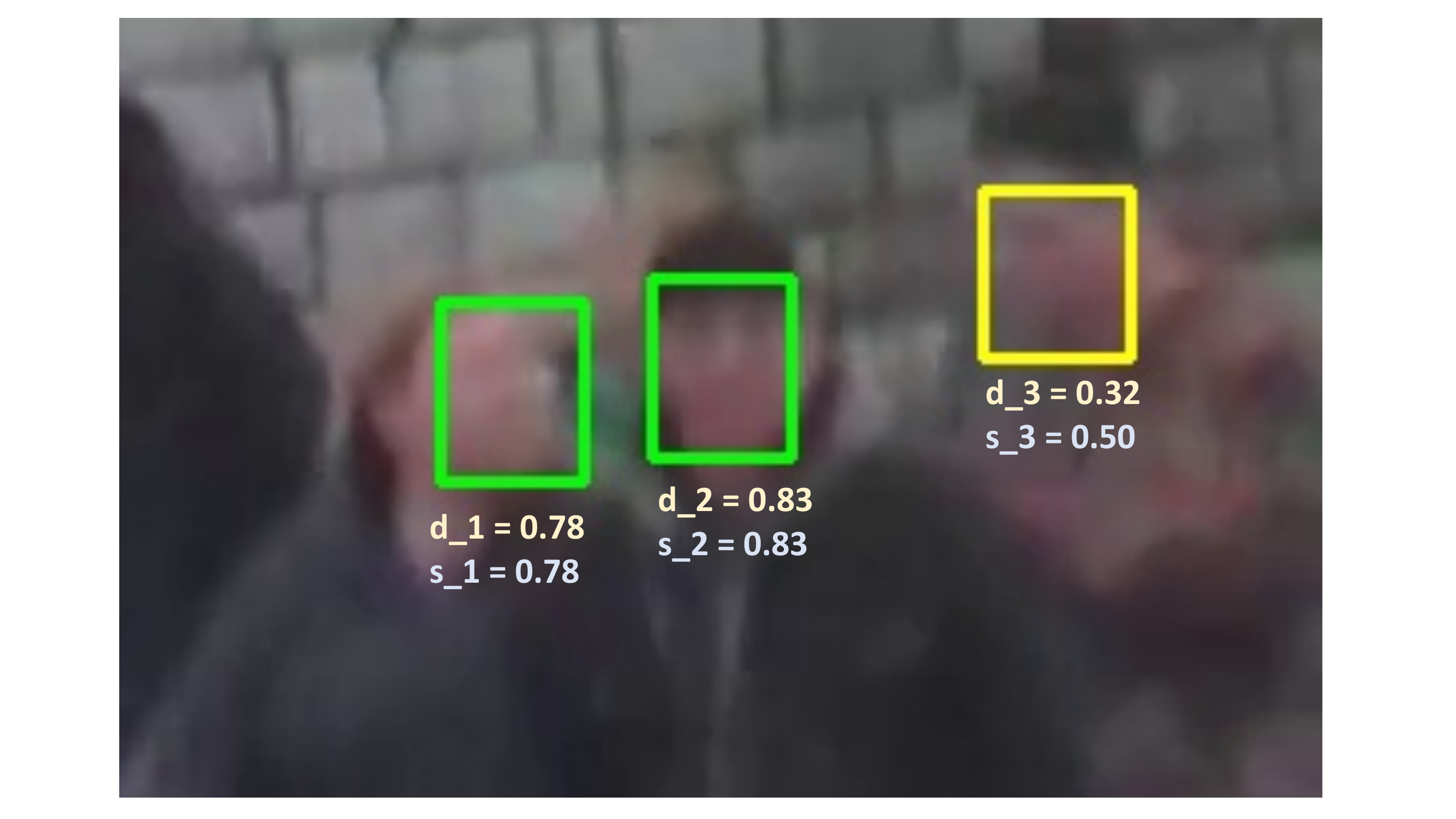}
    \\
    (a)  &  (b) \\
\end{tabular}
\vspace{-4mm}
\caption[]{ \textbf{(a) Pseudo-labels from detection and tracking:}\protect\footnotemark In three consecutive video frames, high-confidence predictions from the baseline detector are marked in \textit{green}, and faces missed by the detector (\ie low detector confidence score) but picked up by the tracker are marked in \textit{yellow}. \textbf{(b) Soft-labeling example:} baseline detector confidences are $d_1=0.78$, $d_2=0.83$, $d_3=0.32$; confidence threshold $\theta=0.5$.   Following Eqn~\ref{eqn:soft-label}, high-confidence detections (\textit{green}) are assigned soft-scores $s_i=d_i$, \ie $s_1=0.78$ and $s_2=0.83$. The tracker-only sample (\textit{yellow}) has detector score below the threshold: $d_3 = 0.32 < \theta$. It gets soft-score $s_3 = \theta = 0.5$. }
\label{fig:tracklet}
\end{figure*}

\noindent
\textbf{Refined labels from tracking.}  
Exploiting the temporal continuity between frames in a video, we can enlarge our set of pseudo-labels with objects missed by the baseline detector. 
To link multiple object detections across video frames into temporally consistent tracklets, we use the algorithm from Jin~\etal(Sec.~3 of \cite{erdosrenyi}) with the MD-Net tracker~\cite{nam2016learning}.
Now, given a tracklet that consistently follows an object through a video sequence, when the object detector did not fire (\ie $d_i < \theta$) in some difficult frames, the tracker can still correctly predict an object (see Fig.~\ref{fig:tracklet}(a)). We expand the set of pseudo-labels to include these ``tracker-only'' bounding-boxes that were missed by the baseline detector, since these \textit{hard examples} are expected to have a larger influence on the model's decision boundary upon retraining~\cite{sung1994learning,shrivastava2016training,jin2018unsup}. Further, we prune out extremely short tracklets (less than 10 frames) to remove the effects caused by spurious detections. 

\footnotetext{Some faces hidden following permissions in \cite{kalkaijb}.}

\label{sec:noisy-train}


\subsection{Training on pseudo-labels}
\label{sec:default_loss}
\noindent
We use the popular Faster R-CNN (FRCNN)~\cite{ren15faster,ren16faster} as our detector. 
In a naive setting, we would treat both labeled source-domain data and pseudo-labeled target-domain data identically in terms of the loss. We give a label of 1 to \textit{all} the target domain pseudo-labeled samples, irrespective of whether it originated from the baseline detector or the tracker -- \ie for $X_i$, the $i$-th training sample drawn from $\mathcal{T}$, the label $y_i$ is defined as
\begin{equation}
  y_i = \begin{cases}
    1, & \text{if $X_i$ is a \textit{pos.} sample (from detector or tracker)}.\\
    0, & \text{if $X_i$ is a \textit{neg.} sample}.
  \end{cases}
\end{equation}
Note that here $X_i$ is not an image, but a \textit{region} in an image.
For training the classification branch, we use a binary cross-entropy loss on the $i$-th training sample:
\begin{equation}
    \mathcal{L}_i(y_i, p_i) = -[y_i \log(p_i) + (1 - y_i) \log(1 - p_i)]
\label{eqn:bce_loss}
\end{equation}

 \noindent 
 where ``\textit{hard}" label $y_i \in \{0,1\}$ and the model's predicted posterior $p_i \in [0,1]$. This is similar to the method of Jin~\etal~\cite{jin2018unsup}, which assigns a label of 1 for both easy and hard positive examples during re-training.

\subsection{Distillation loss with soft labels}
\label{sec:distill}
\noindent
For training data coming from $\mathcal{T}$, many of the $y_i$s can be noisy, so a ``\textit{soft}" version of the earlier $\{0,1\}$ labels could help mitigate the risk from mislabeled target data.
Label smoothing in this fashion has been shown to be useful in generalization~\cite{szegedy2016rethinking,hinton2015distilling}, in reducing the negative impact of incorrect training labels~\cite{li2017learning} and is more informative about the distribution of labels than one-hot encodings~\cite{tzeng2015simultaneous}.  
In our case, each target-domain \textit{positive} label can have two possible origins -- (i) high-confidence predictions from the baseline detector or (ii) the tracklet-formation process. 
We assign a \textbf{\textit{soft score}} $s_i$ to each positive target-domain sample $X_i \in \mathcal{T}$ as follows:
\begin{equation}
  s_i=\begin{cases}
    d_i, & \text{if $X_i$ originates from  detector}.\\
    \theta, & \text{if $X_i$ originates from tracker}.
  \end{cases}
  \label{eqn:soft-label}
\end{equation}
For a pseudo-label originating from the baseline detector, a high detector confidence score $d_i$ is a reasonable measure of reliability. Tracker-only pseudo-labels, which could be objects missed by the baseline model, are emphasized during training -- their soft score is raised up to the threshold $\theta$, although the baseline's confidence on them had fallen below this threshold. An illustrative example is shown in Fig.~\ref{fig:tracklet}(b).

\noindent
\textbf{Label interpolation.} 
A \textbf{\textit{soft label}} $\tilde{y}_i$ is formed by a linear interpolation between the earlier hard labels $y_i$ and soft scores $s_i$, with $\lambda \in [0,1]$ as a tunable hyper-parameter. 
\begin{equation}
    \tilde{y}_i =  \lambda s_i + (1 - \lambda) y_i
    \label{eqn:tilde}
\end{equation}

\noindent
The loss for the $i$-th positive sample now looks like
\begin{equation}
  \mathcal{L}_i^{distill} = 
    \begin{cases}
        \mathcal{L}_i(y_i, p_i),          & \text{if $X_i \in \mathcal{S}$}.\\
        \mathcal{L}_i(\tilde{y}_i, p_i),  & \text{if $X_i \in \mathcal{T}$}.
    \end{cases}
\end{equation}
Setting a high value of $\lambda$ creates softer labels $\tilde{y}_i$, trusting the baseline source model's prediction $s_i$ more than than the riskier target pseudo-labels $y_i$. In this conservative setting, the softer labels will decrease the overall training signal from target data, but also reduces the chance of incorrect pseudo-labels having a large detrimental effect on the model parameters. 

\noindent
We now describe two schemes to avoid explicitly depending on the $\lambda$ hyper-parameter --

\noindent
\textbf{I. Constrained hard examples.} 
\label{sec:blah}
Assigning a label of 1 to both ``easy'' and ``hard'' examples (\ie high-confidence detections and tracker-only samples), as in Sec.~\ref{sec:default_loss}, gives equal importance to both. Training with \textit{just} the hard examples can be sub-optimal -- it might decrease the model's posteriors on instances it was getting correct initially. Ideally, we would like to emphasize the hard examples, while simultaneously \textit{constraining} the model to maintain its posteriors on the other (easy) samples. 
We can achieve this by setting $\theta=1$ in Eq.~\ref{eqn:soft-label} and $\lambda=1$ in Eq.~\ref{eqn:tilde}, which would create a label of 1 for tracker-only ``hard'' examples, and a label equal to baseline detector score for the high-confidence detections, \ie ``easy'' examples.

\begin{figure}
\centering
\tabcolsep=0.08cm
\begin{tabular}{ccc}
    \includegraphics[width=0.16\textwidth]{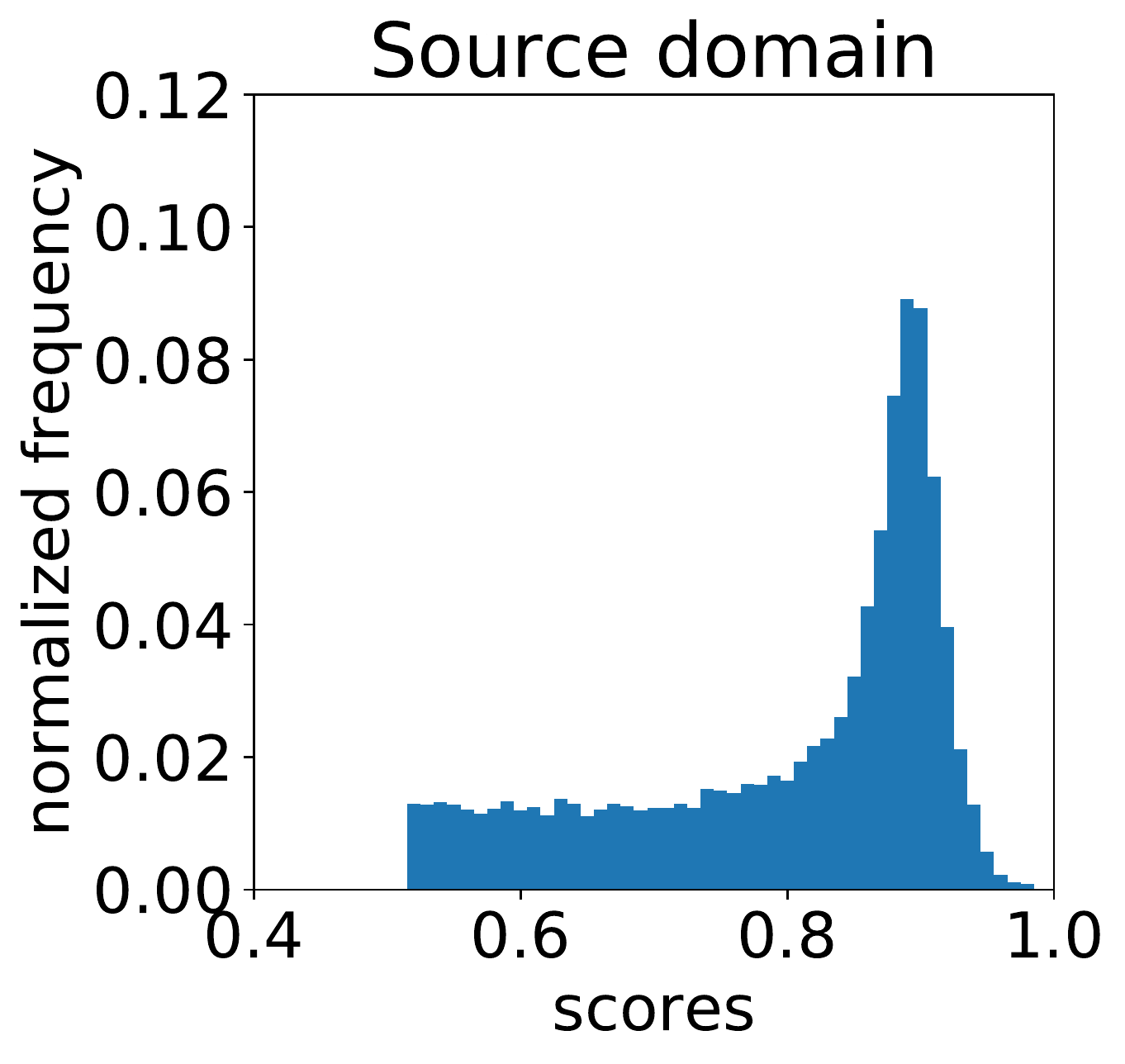}
    &
    \includegraphics[width=0.135\textwidth]{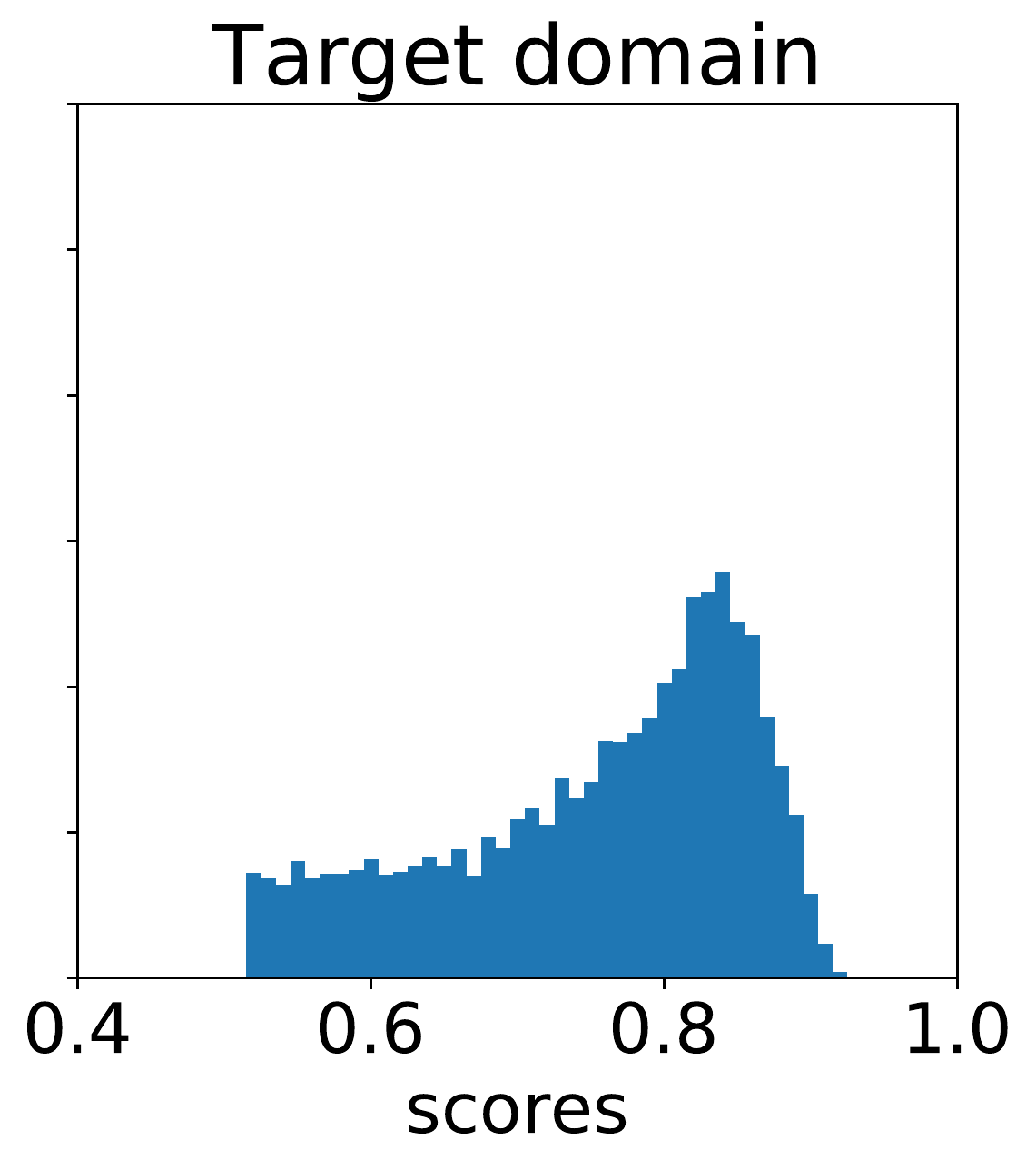}
    &
    \includegraphics[width=0.145\textwidth]{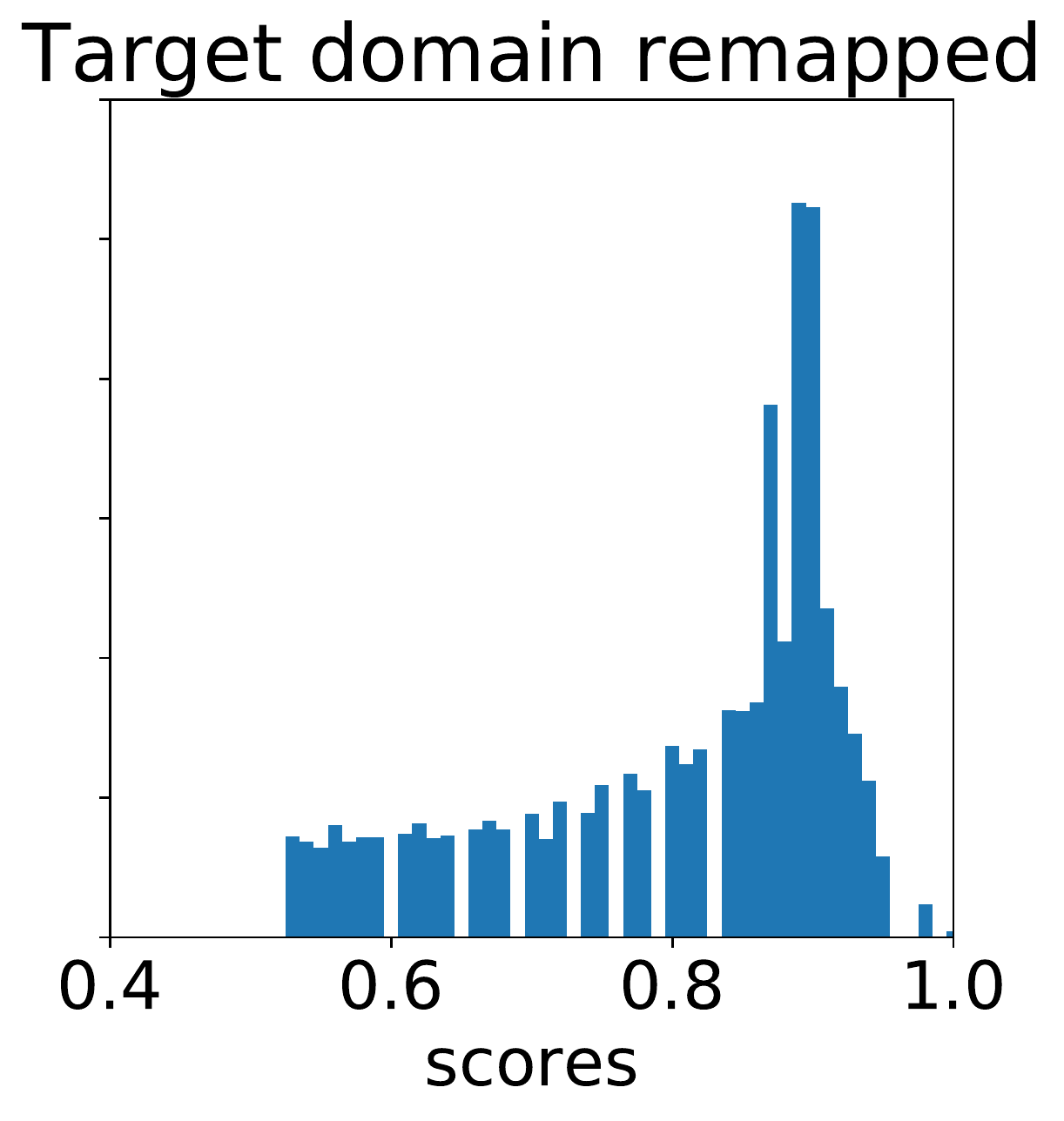}
    \\
    (a)  &  (b) & (c) \\
\end{tabular}
\vspace{-4mm}
\caption{\textbf{Cross-domain score mapping.} Distribution of high-confidence detection scores of a face detector trained on labeled images from WIDER-Face~\cite{yang16wider}; samples are from (a) WIDER-validation and (b) CS6 surveillance videos~\cite{kalkaijb}; (c) remapping the scores on CS6 to resemble WIDER. 
}
\vspace{-4mm}
\label{fig:score_hist}
\end{figure}

\noindent
\textbf{II. Cross-domain score mapping.} 
\label{sec:score_map}
Let us hypothetically consider what the distribution of detection scores on $\mathcal{T}$ would be like, had the model been trained on \textit{labeled} target domain data. With minimal information on $\mathcal{T}$, it is reasonable to assume this distribution of scores to be similar to that on $\mathcal{S}$. The latter is an ``ideal'' operating condition of training on labeled data and running inference on within-domain images. 
Let the actual distribution of baseline detector scores on $\mathcal{T}$ have p.d.f. $f(x)$, and the distribution of scores on $\mathcal{S}$ have p.d.f. $g(x)$. Let their cumulative distributions be $F(x) = \int_{0}^{x} f(t) dt$ and $G(x) = \int_{0}^{x} g(r) dr$, respectively. As a parameter-free method of creating soft-labels for our pseudo-labels on $\mathcal{T}$, we can use histogram specification~\cite{gonzalez2002digital} to map the baseline detector scores on $\mathcal{T}$ to match the distribution of scores on images from $\mathcal{S}$, \ie replace each target domain score $x$ with $G^{-1}(F(x))$. The inverse mapping is done through linear interpolation.
Fig.~\ref{fig:score_hist}(a) shows the distribution of scores for a model trained on labeled WIDER-Face~\cite{yang16wider} and run on images from the validation split of the same dataset. In Fig.~\ref{fig:score_hist}(b), due to the domain shift, there is a visible difference when this model is run on unlabeled images from CS6 surveillance videos~\cite{kalkaijb}. Fig.~\ref{fig:score_hist}(c) shows the effect of histogram matching. 
Concretely, detector samples get soft-label $G^{-1}(F(d_i))$, while tracker-only samples get soft-label $\theta$.

\section{Experiments}
\label{sec:experiments}

\noindent
The datasets are introduced in Sec.~\ref{sec:datasets}, followed by describing baselines (Sec.~\ref{sec:ablation}) and implementation details (Sec.~\ref{sec:eval}). Results are shown on faces (Sec.~\ref{sec:face_result}) and pedestrians (Sec.~\ref{sec:ped_result}).

\subsection{Datasets}
\label{sec:datasets}
\noindent
Experiments are performed on two challenging scenarios -- pedestrian detection from driving videos and face detection from surveillance videos, both of which fit neatly into our paradigm of self-training from large amounts of unlabeled videos and where there exists a significant domain shift between source and target. Several example images are shown in Fig.~\ref{fig:task_example}. We select single-category detection tasks like face and pedestrian to avoid the engineering and computational burden of handling multiple categories, and focus on the unsupervised domain adaptation aspect. A summary of the datasets is given in Table~\ref{tab:data_stats}.

\noindent
\textbf{Face: WIDER $\rightarrow$ CS6.} 
The WIDER dataset~\cite{yang16wider} is the the source domain, consisting of labeled faces in still images downloaded from the internet with a wide variety of scale, pose and occlusion. The baseline detector is trained on the WIDER Train split, which has 12,880 images and 159,424 annotated faces. 
The target domain consists of 179 surveillance videos from CS6, which is a subset of the IJB-S benchmark~\cite{kalkaijb}. CS6 provides a considerable shift from WIDER, with faces being mostly low-resolution and often occluded, and the imagery being of low picture quality, suffering from camera shake and motion blurs. The video clips are on average of 5 minutes at 30 fps, with some exceptionally long clips running for over an hour. 
We selected 86 videos to form the unlabeled target train set (\textit{CS6-Train}). A test set of 80 labeled videos, containing about 70,022 images and 217,827 face annotations, is used to evaluate the performance of the methods (\textit{CS6-Test}).

\noindent
\textbf{Pedestrian: BDD(\textit{clear,daytime}) $\rightarrow$ BDD(\textit{rest}).}
The Berkeley Deep Drive 100k (BDD-100k) dataset~\cite{yu2018bdd100k} consists of 100,000 driving videos from a wide variety of scenes, weather conditions and time of day, creating a challenging and realistic scenario for domain adaptation. Each video clip is of 40 seconds duration at 30 fps; one frame out of every video is manually annotated. 
The source domain consists of clear, daytime conditions (\textit{BDD(clear,daytime)}) and the target domain consists of all other conditions including night-time, rainy, cloudy, \etc (\textit{BDD(rest)}).
There are 12,477 labeled images forming \textit{BDD-Source-Train}, containing 217k pedestrian annotations. We use 18k videos as the unlabeled \textit{BDD-Target-Train} set, having approximately 21.6 million video frames (not all of which would contain pedestrians, naturally). The \textit{BDD-Target-Test} set is comprised of 8,236 labeled images with 16,784 pedestrian annotations from \textit{BDD(rest)}.

\begin{table}
\renewcommand{\tabcolsep}{5pt}
\centering
\small 
\caption{
    \textbf{Dataset summary.} Details of the source and target datasets for face and pedestrian detection tasks are summarized here. N.B.-- for the unlabeled target train sets, the \#images and \#annotations are unknown.} 
\vspace{-3mm}
\label{tab:data_stats} 
\begin{tabular}{@{\extracolsep{5pt}}lccc}
\toprule
 \textbf{Dataset}            & \textbf{\# images}      & \textbf{\# annots}    & \textbf{\# videos} \\
\midrule
WIDER                     & 12,880                  & 159,424               &   -  \\
CS6-Train                 & -                       & -                     &   86  \\
CS6-Test                  & 70,022                  & 217,827               &   80  \\
\midrule
BDD-Source                       & 12,477                    & 16,784                &   12,477  \\
BDD-Target-Train                 & -                         & -                     &   18,000  \\
BDD-Target-Test                  & 8,236                     & 10,814                &   8,236  \\
\bottomrule
\end{tabular}
\vspace{-4mm}
\end{table}

\subsection{Baselines and Ablations}
\label{sec:ablation}
\noindent
We consider the following methods as our baselines:

\noindent
\textbf{Baseline source.} Detector trained on only the labeled source data -- WIDER for faces and BDD(\textit{clear,daytime}) for pedestrians. 
    
\noindent
\textbf{Pseudo-labels from detections.} 
  High-confidence detections on the target training set are considered as training labels, followed by joint re-training of the baseline source detector. This is the naive baseline for acquiring pseudo-labels, denoted as \texttt{Det} in the results tables.
  
\noindent
\textbf{Pseudo-labels from tracking.} Incorporating temporal consistency using a tracker and adding them into the set of pseudo-labels was referred to as \textit{``Hard Positives''} by Jin~\etal~\cite{jin2018unsup}; we adopt their nomenclature and refer to this as \texttt{HP}. As an ablation, we exclude detector results and keep just the \textit{tracker-only} pseudo-labels for training (\texttt{Track}). Table~\ref{tab:mining_stats} summarizes the details of the automatically gathered pseudo-labels. Note that using temporal constraints (\texttt{HP}) removes spurious isolated detections in addition to adding missed objects, resulting in an overall decrease in data when compared to \textit{Det} for CS6.

\noindent
\textbf{Soft labels for distillation.} The \textit{label interpolation} method as detailed in Sec.~\ref{sec:distill} is denoted as \texttt{Label-smooth}, and we show the effect of varying $\lambda$ on the validation set. Cross-domain score distribution mapping is referred to as \texttt{score-remap} and constrained hard examples as \texttt{HP}\textit{-cons} in the results tables.

\noindent
\textbf{Domain adversarial Faster-RCNN.}  
While there are several domain adversarial methods such as ADDA~\cite{tzeng2017adversarial} and CyCADA~\cite{hoffman2017cycada} for object \textit{recognition}, we select Chen~\etal~\cite{chen2018domain} as the only method, to our knowledge, that has been integrated into the Faster R-CNN \textit{detector}. Chen~\etal~\cite{chen2018domain} formulate the adversarial domain discriminator~\cite{ganin2014unsupervised} with three separate losses -- (i) predicting the domain label from the convolutional features (pre-ROI-pooling) of the entire image; (ii) predicting the domain label from the feature-representation of each proposed ROI; (iii) a consistency term between the image-level and ROI-level predictions. The region-proposals for the ROI-level loss are obtained from the Region Proposal Network (RPN) branch of the Faster R-CNN. In our experiments, we denote these models as -- \texttt{DA}\textit{-im} which applies the domain discriminator at the image level and \texttt{DA}\textit{-im-roi}, which additionally has the instance-level discriminator and consistency term.

\begin{table}
\renewcommand{\tabcolsep}{5pt}
\centering
\small 
\caption{
    \textbf{Pseudo-labels summary.} Listing the number of images and object annotations obtained on the unlabeled \textit{CS6-Train} and \textit{BDD-Target-Train} videos. All the pseudo-labels obtained from the CS6 videos are used in re-training. For BDD, due to the massive number of videos, 100K frames were sub-sampled to form the training set.  
}
\vspace{-3mm}
\label{tab:mining_stats} 
\begin{tabular}{@{\extracolsep{5pt}}lcc}
\toprule
\textbf{Method}          & \textbf{\# images}  & \textbf{\# annots}     \\
\midrule
CS6-Det                   &  38,514            &  109,314    \\
CS6-HP                    &  15,092            &  84,662     \\
CS6-Track                 &  15,092            &  32,711     \\
\midrule                
BDD-Det                 & 100,001          &  205,336   \\
BDD-Track               &  100,001         &   222,755    \\
BDD-HP                  &  100,001         &   362,936   \\
\bottomrule
\label{tab:mining_stats}
\vspace{-1cm}
\end{tabular}
\end{table}

\subsection{Training and Evaluation}
\label{sec:eval}
\noindent
We use the standard Faster R-CNN detector~\cite{ren15faster} for all our experiments, from a PyTorch implementation of the Detectron framework~\cite{Detectron2018}. An ImageNet-pre-trained ResNet-50 network is used as a backbone, with ROI-Align region pooling. For faces, the baseline is trained for 80k iterations, starting from a learning rate of 0.001, dropping to 0.0001 at 50k, using 4 GPUs and a batch-size of 512. For pedestrians, the baseline was trained for 70k iterations, starting with a learning rate of 0.001 and dropping to 0.0001 at 50k. During training, face images were resized to be 800 pixels and pedestrian images were resized to be 500 pixels on the smaller side.
Re-training for the target domain is always done jointly, using a single GPU -- a training mini-batch is formed with samples from a labeled source image and a pseudo-labeled target image. In practice, we sample images alternately from source and target, fix 64 regions to be sampled from each image, and accumulate gradients over the two images before updating the model parameters. 
Domain adversarial models were implemented following Chen~\etal~\cite{chen2018domain}, keeping their default hyper-parameter values.

\noindent
Since unsupervised learning considers no labels \textit{at all} on the target domain, we cannot set hyper-parameters or do best model selection based on a labeled validation set. The re-training for \textit{all} the face models were stopped at the 10k iteration, while \textit{all} the pedestrian models were stopped at the 30k iteration. 
For evaluating performance, to account for stochasticity in the training procedure, we do 5 rounds of training and evaluate each model on the labeled images from the test set.
We use the MS-COCO toolkit as a consistent evaluation metric for both face and pedestrian detection, reporting Average Precision (AP) at an IoU threshold of 0.5.

\begin{figure*}
\centering
\small 
\setlength{\tabcolsep}{1pt} 
\renewcommand{\arraystretch}{0.5} 
\begin{tabular}{cccccc}
    & 1 & 2 & 3 & 4 & 5 \\
    a.
        &
        \includegraphics[trim=15in 5in 6in 7in,clip=true,width=0.18\textwidth]{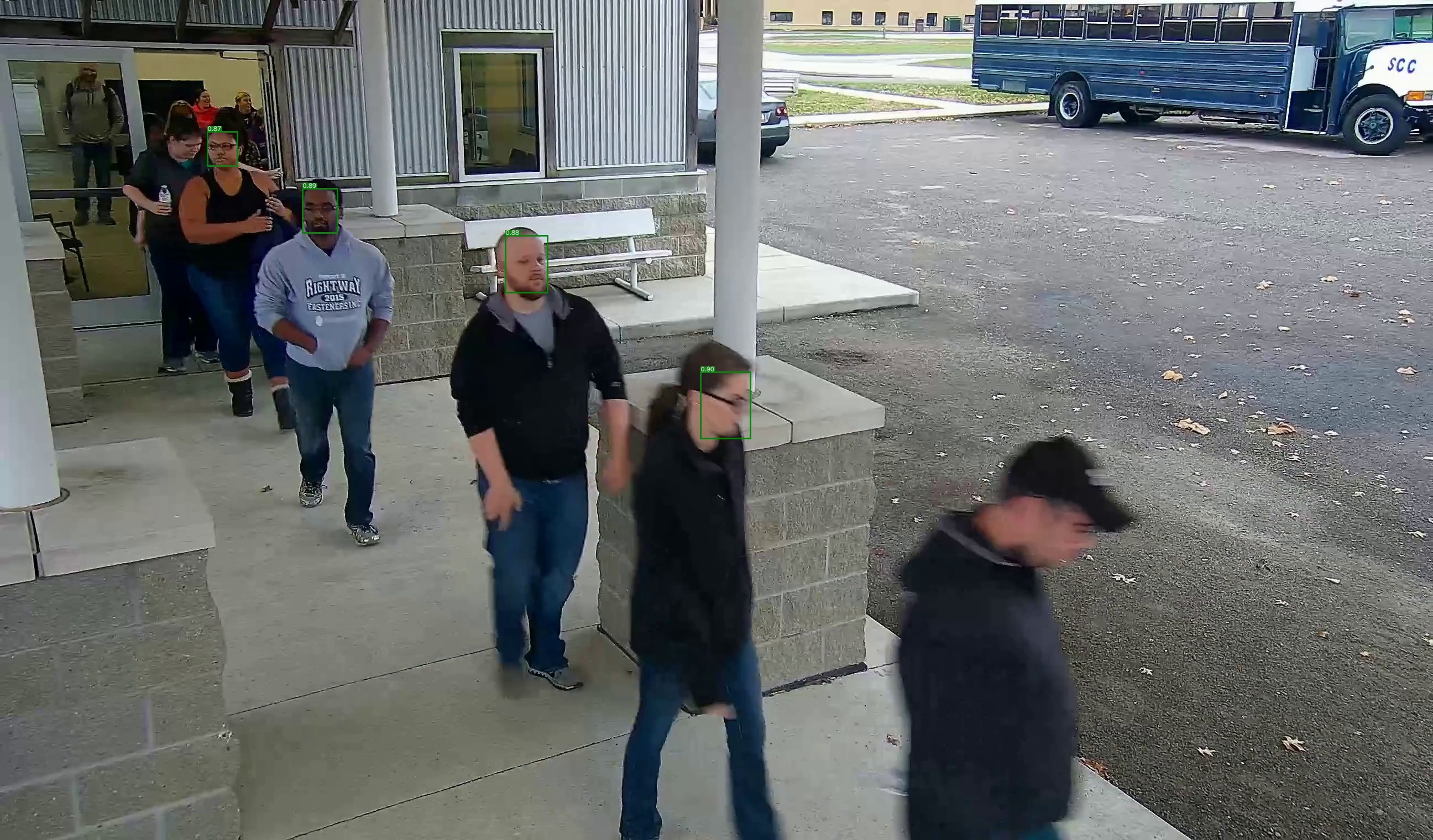}
        &
        \includegraphics[trim=0 6in 10in 2in,clip=true,width=0.16\textwidth]{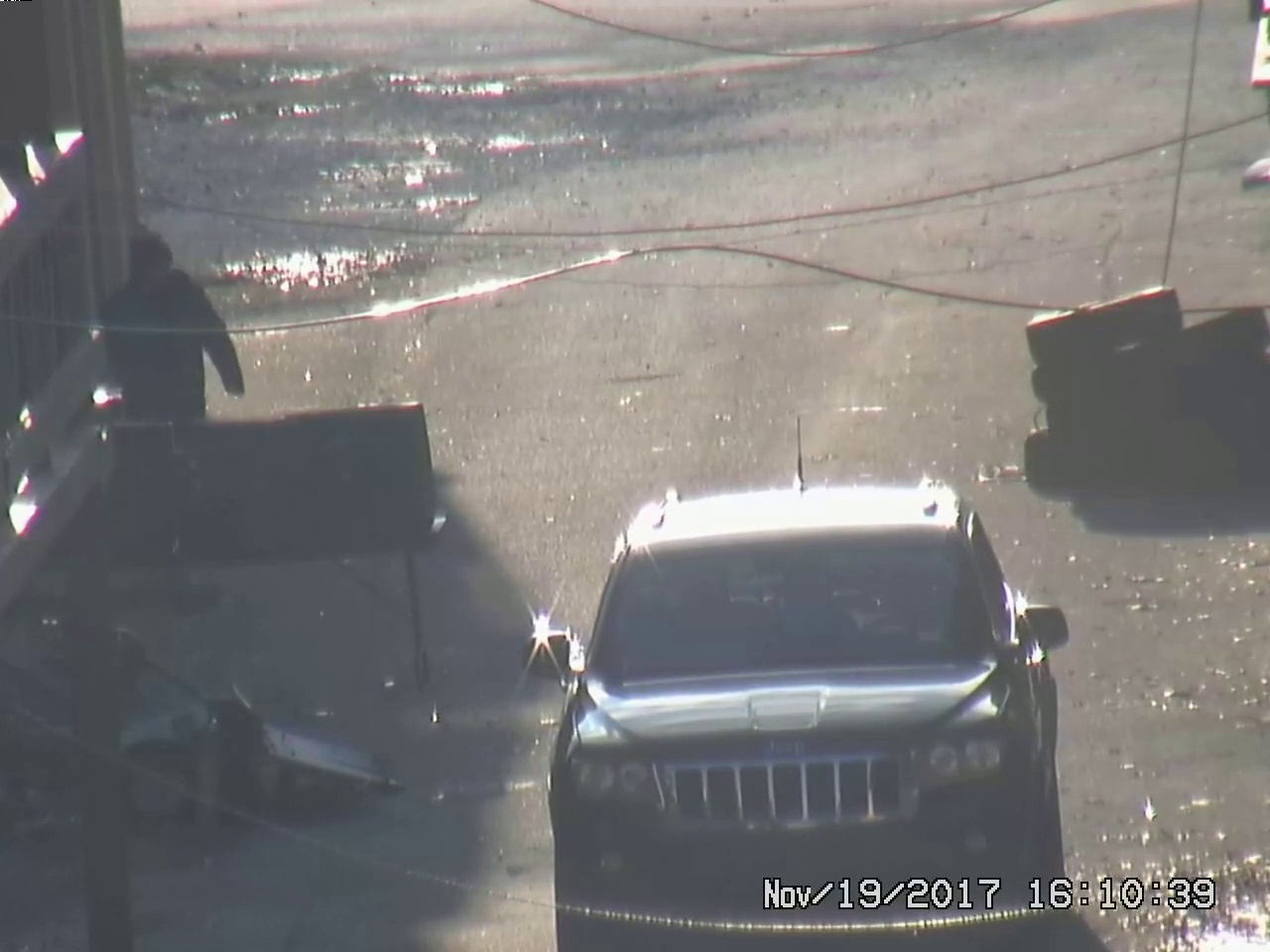}  
        &
        \includegraphics[width=0.19\textwidth]{figures/bdd/images/baseline/b242f6b2-f5da110f.jpg} 
        &
        \includegraphics[width=0.19\textwidth]{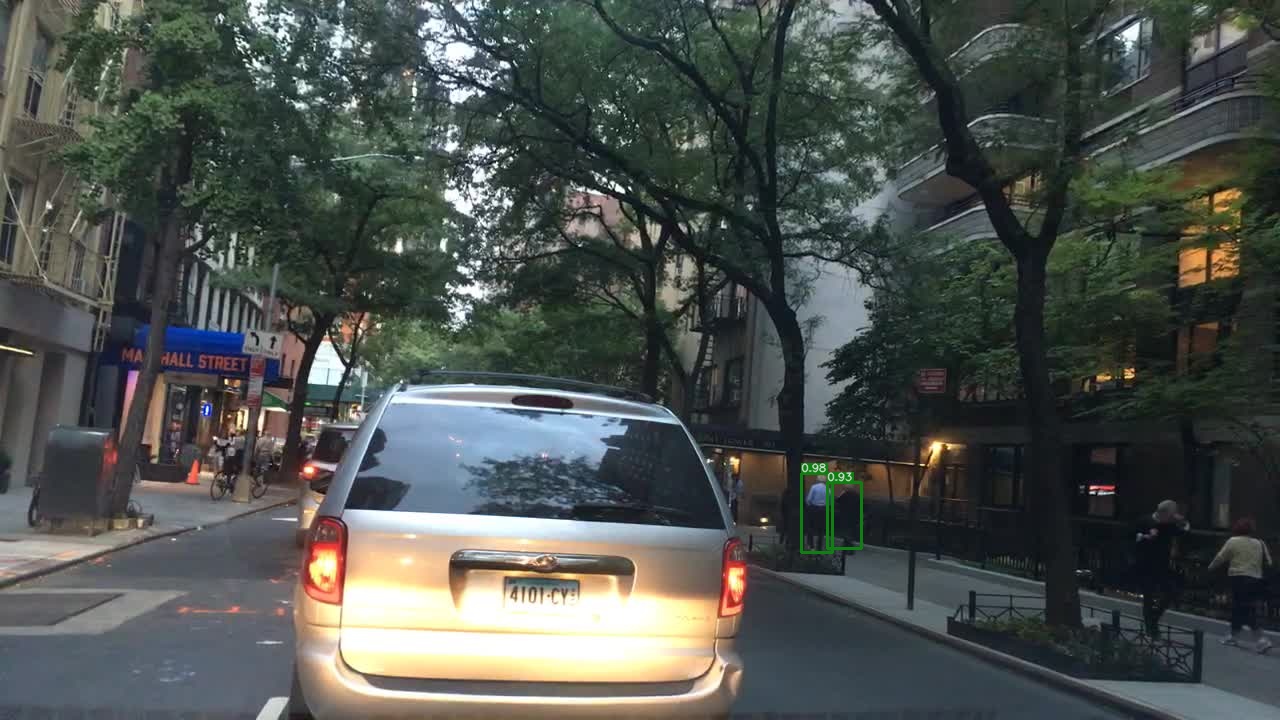} 
        &
        \includegraphics[width=0.19\textwidth]{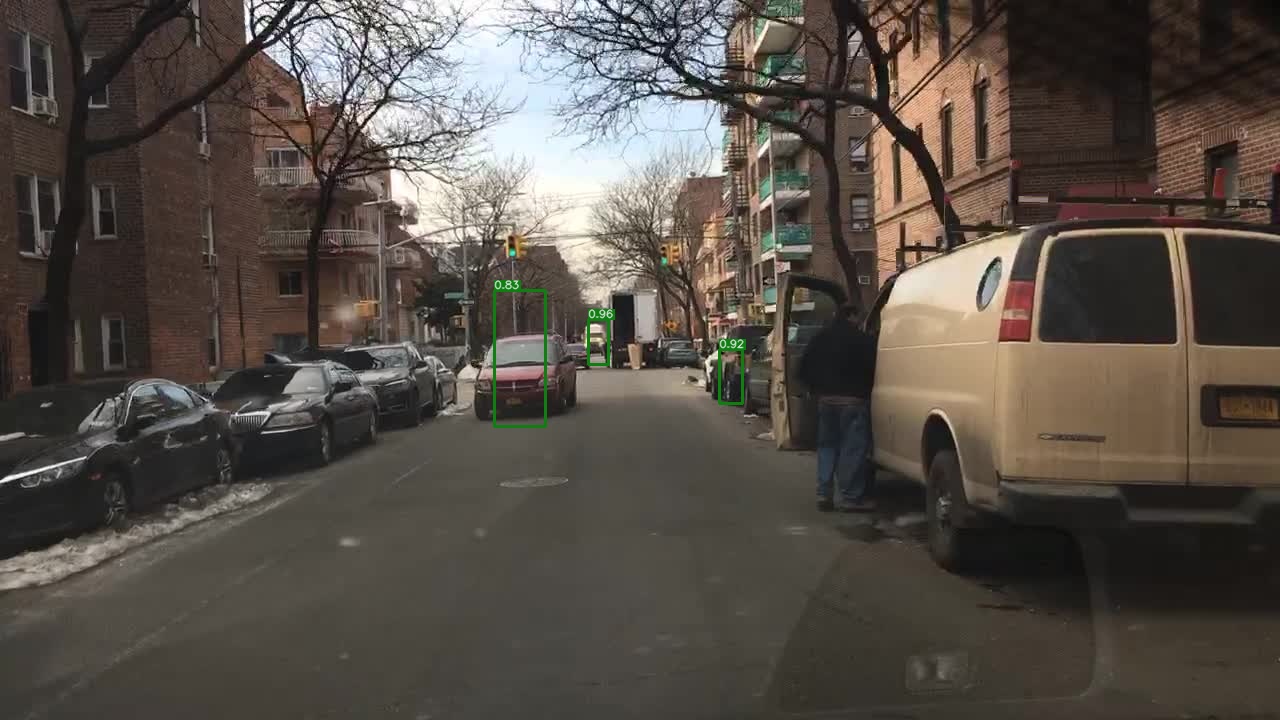} 
    \\
    b.
        &
        \includegraphics[trim=15in 5in 6in 7in,clip=true,width=0.18\textwidth]{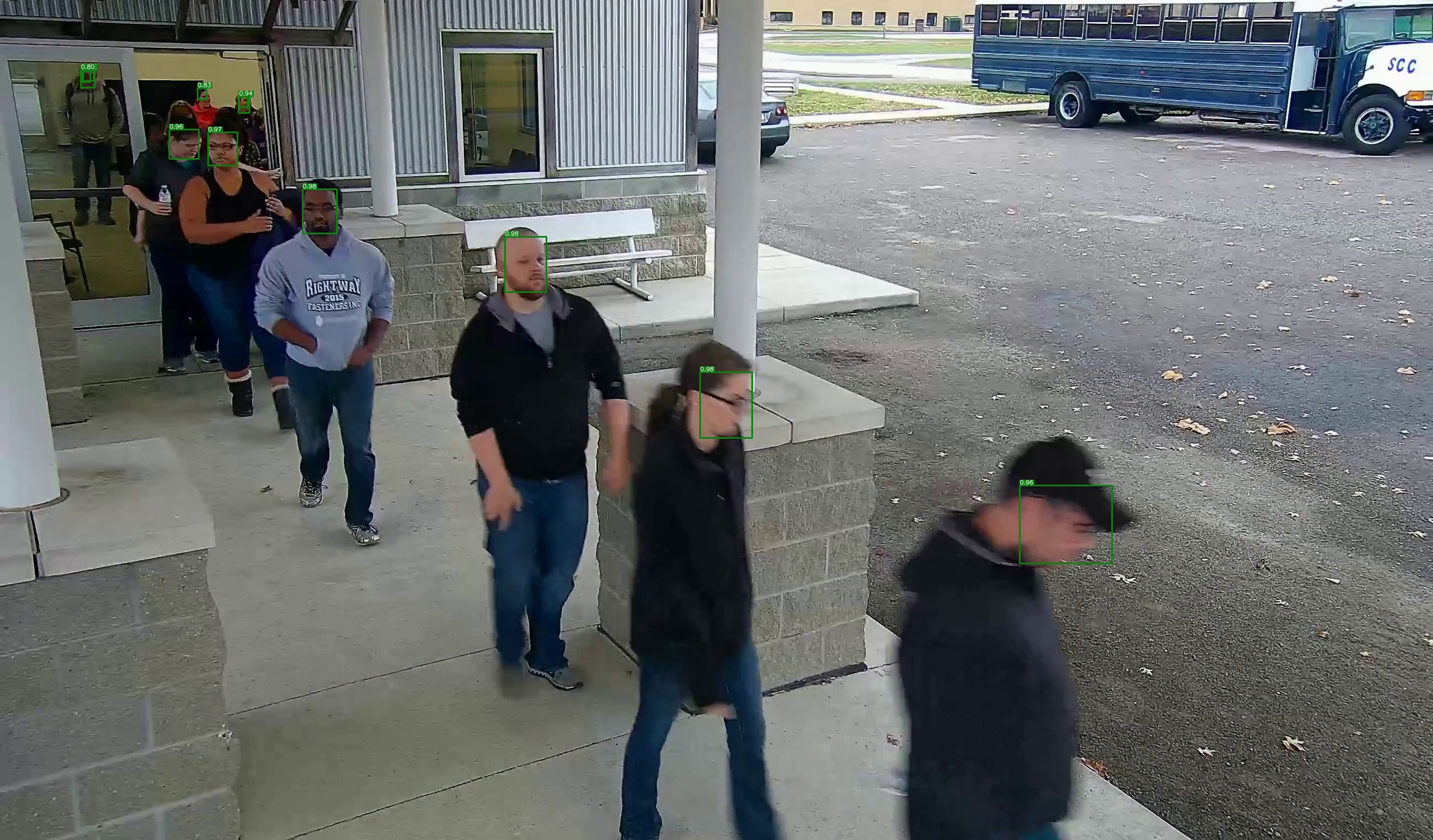}
        &
        \includegraphics[trim=0 6in 10in 2in,clip=true,width=0.16\textwidth]{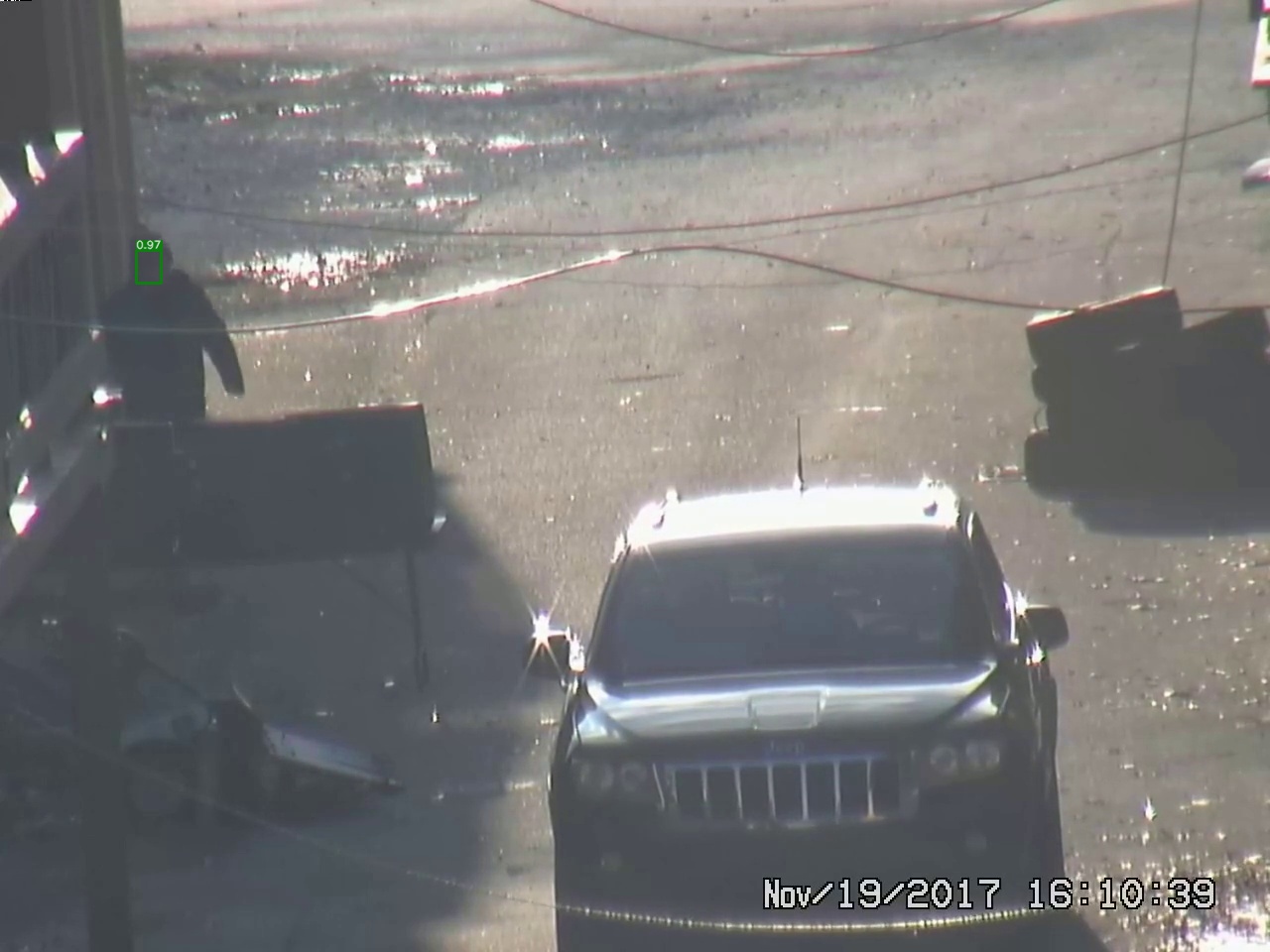} 
        &
        \includegraphics[width=0.19\textwidth]{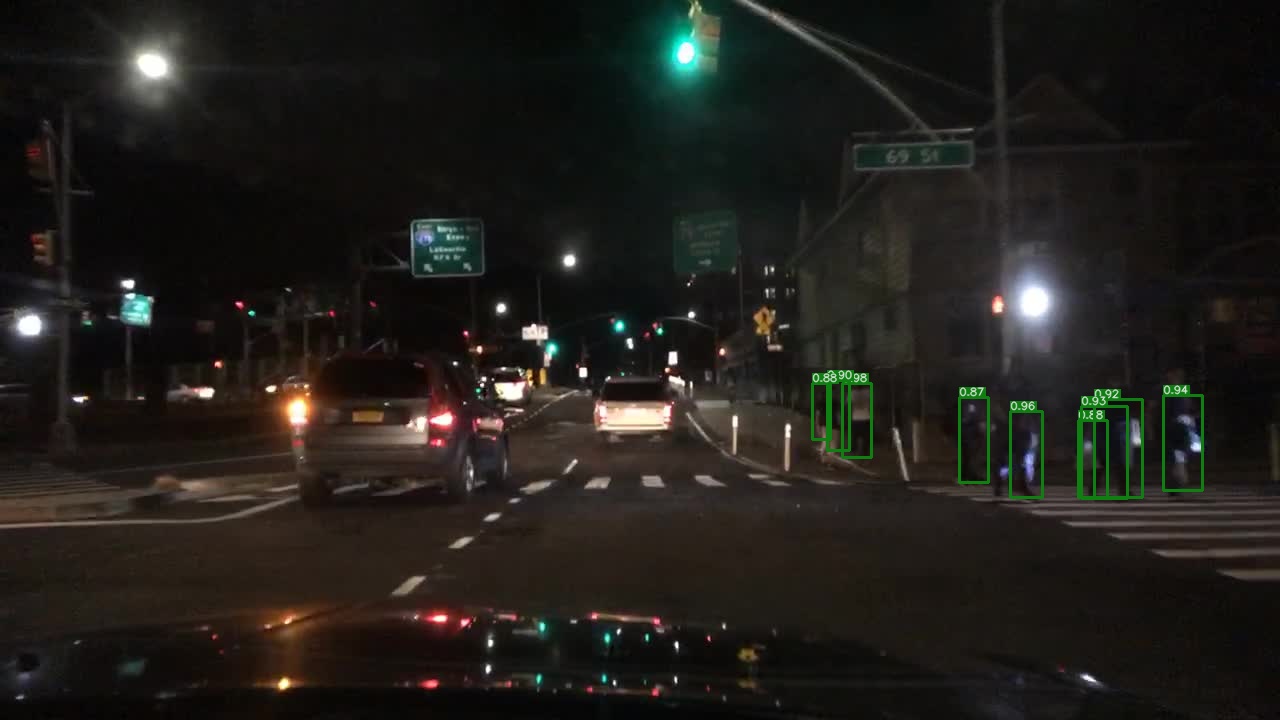} 
        &
        \includegraphics[width=0.19\textwidth]{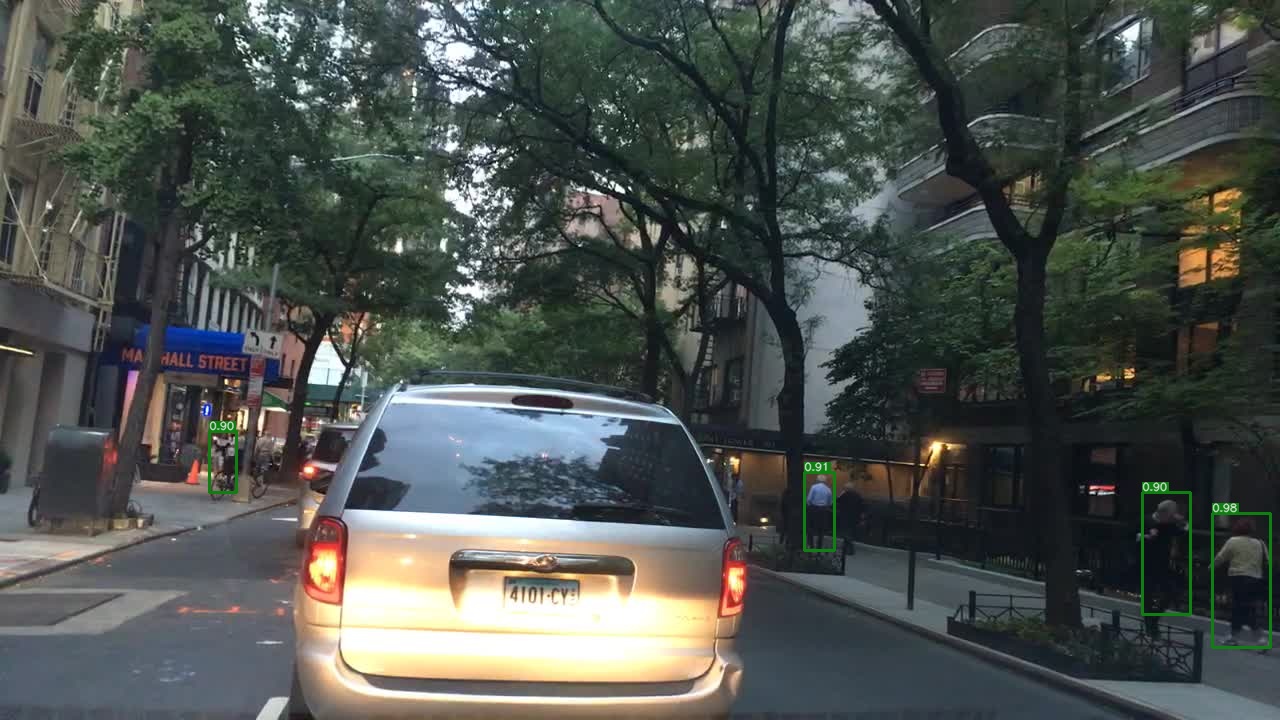} 
        &
        \includegraphics[width=0.19\textwidth]{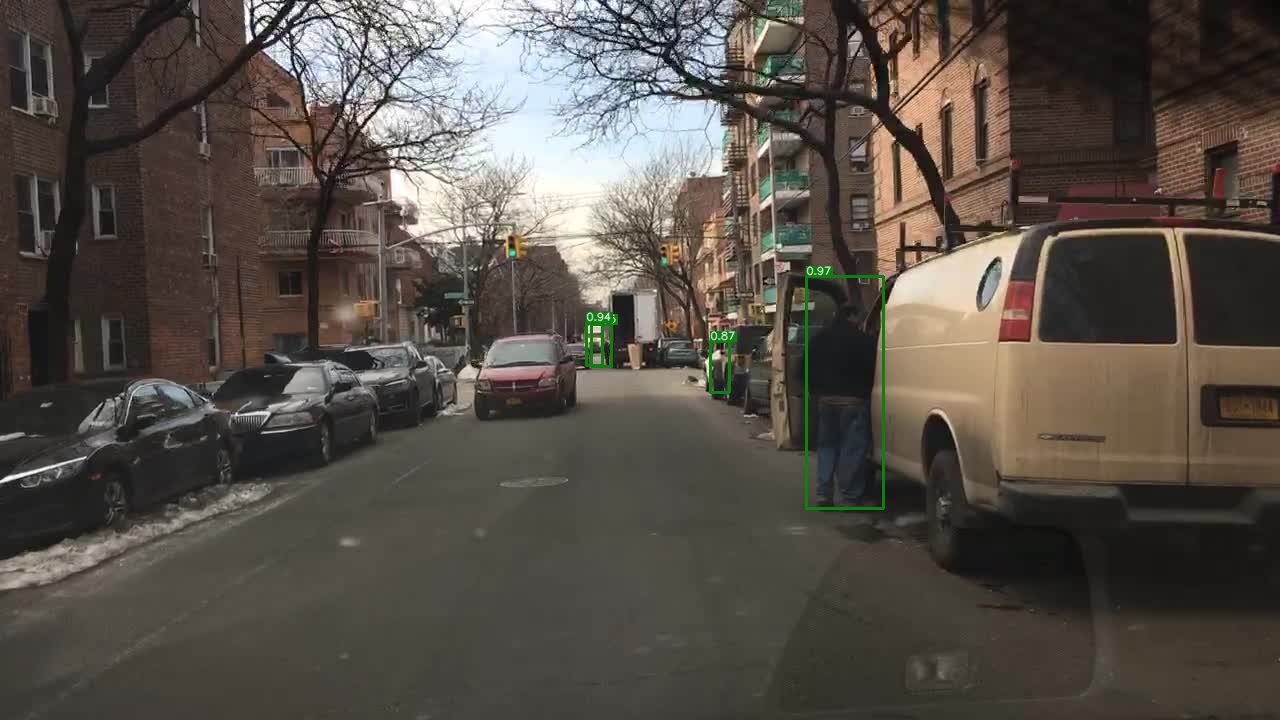}  
    \\
    c. 
        &
        \includegraphics[trim=15in 5in 6in 7in,clip=true,width=0.18\textwidth]{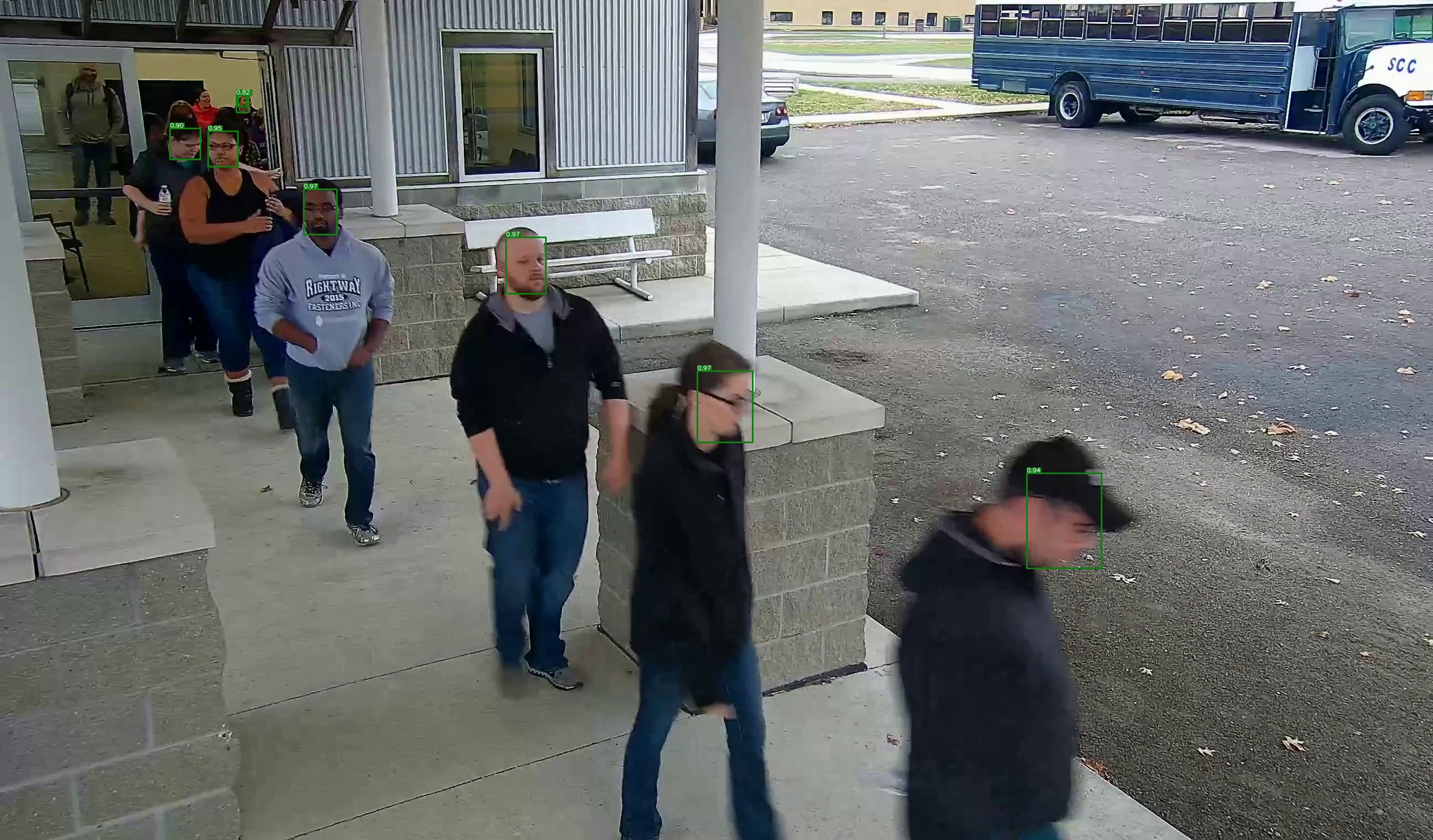}
        &
        \includegraphics[trim=0 6in 10in 2in,clip=true,width=0.16\textwidth]{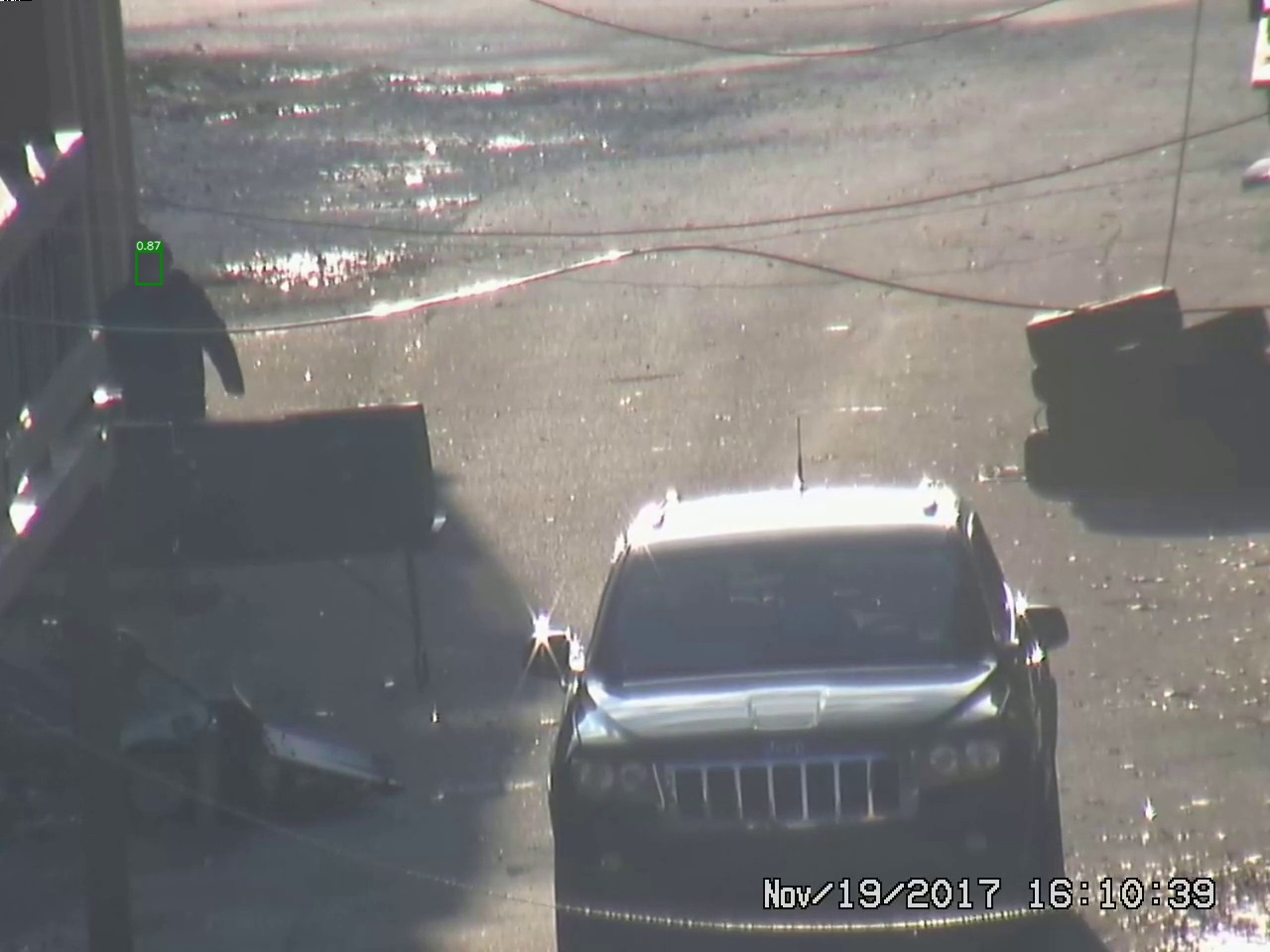}  
        &
        \includegraphics[width=0.19\textwidth]{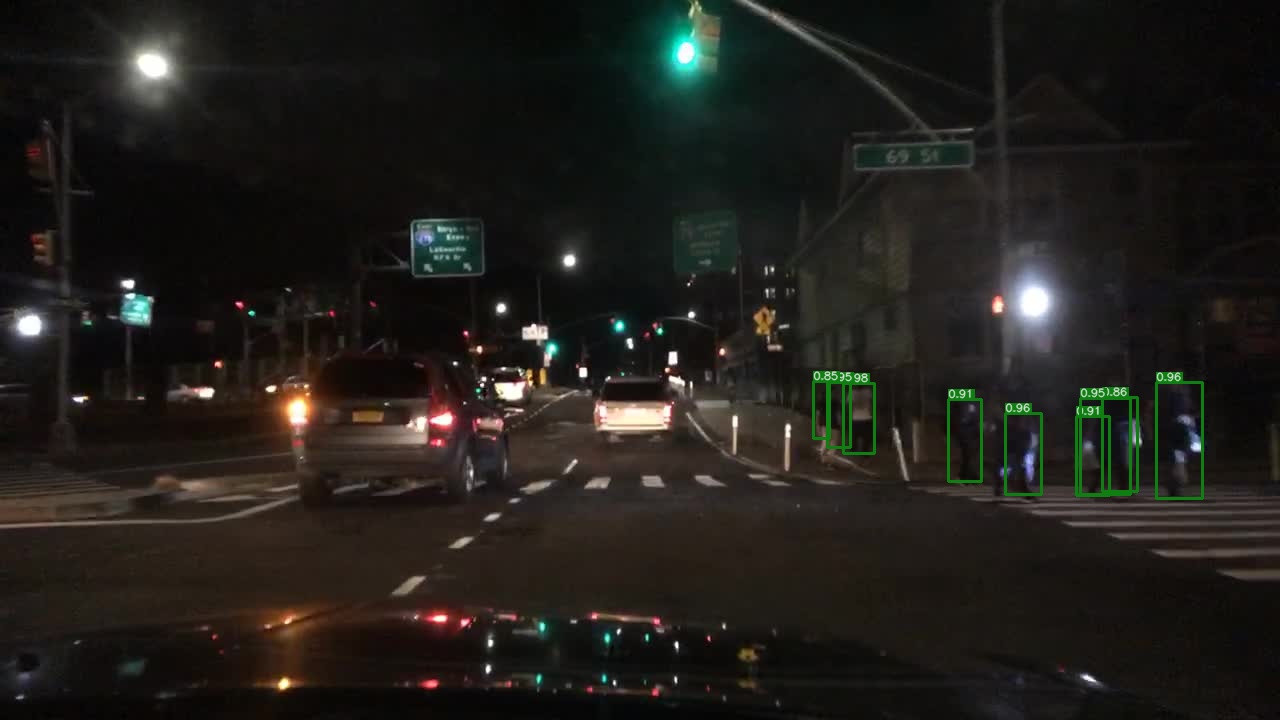} 
        &
        \includegraphics[width=0.19\textwidth]{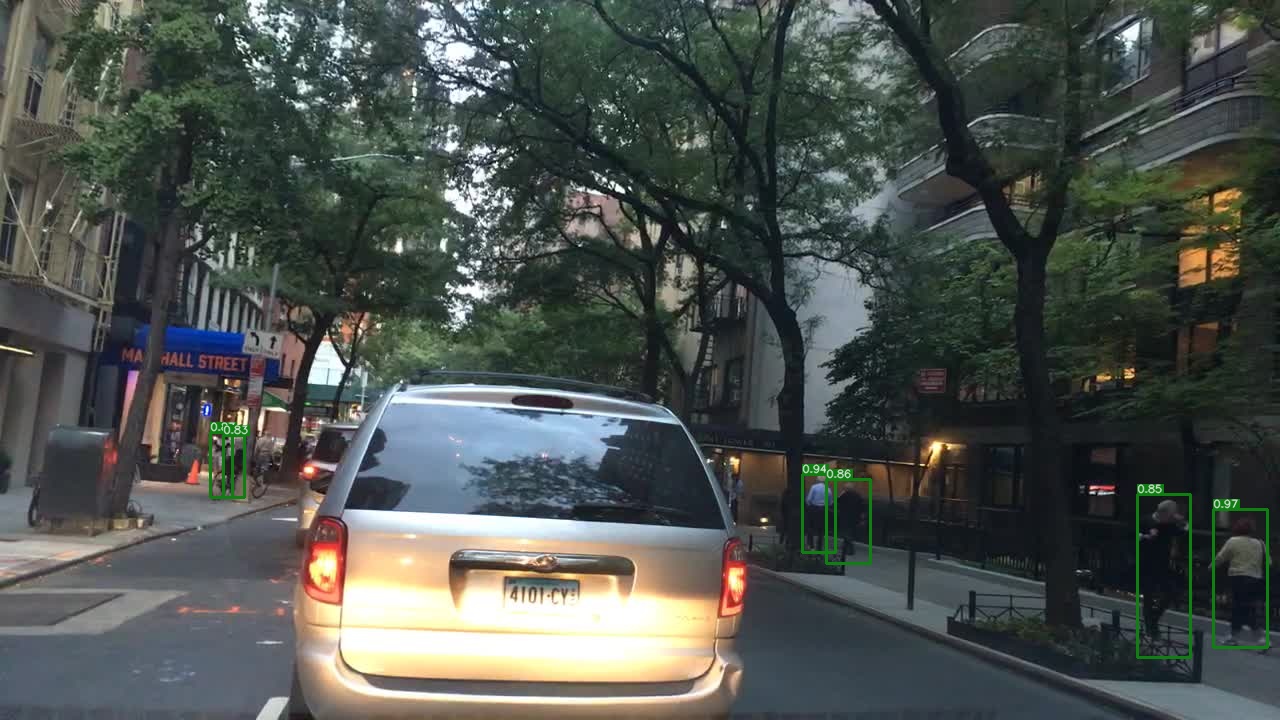} 
        &
        \includegraphics[width=0.19\textwidth]{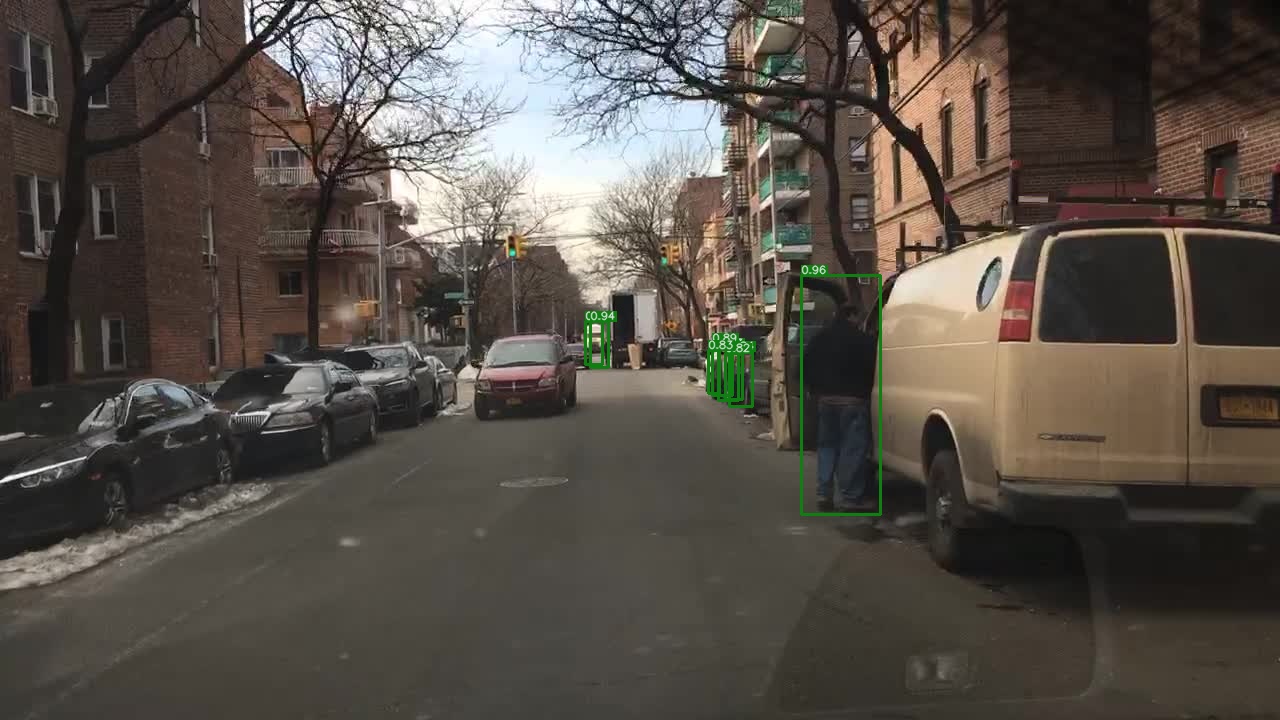}  
    \\
    d. 
        &
        \includegraphics[trim=15in 5in 6in 7in,clip=true,width=0.18\textwidth]{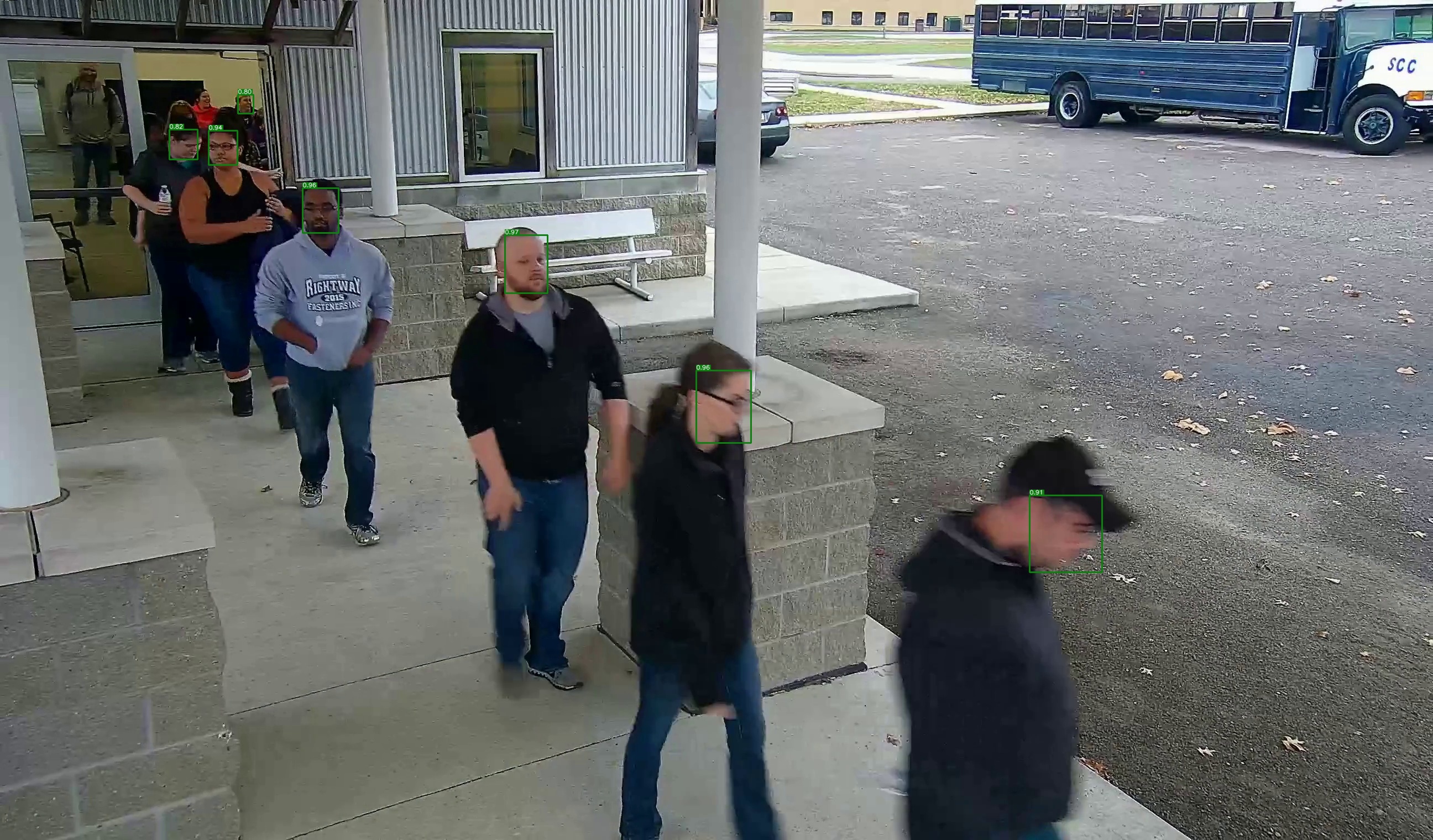}
        &
        \includegraphics[trim=0 6in 10in 2in,clip=true,width=0.16\textwidth]{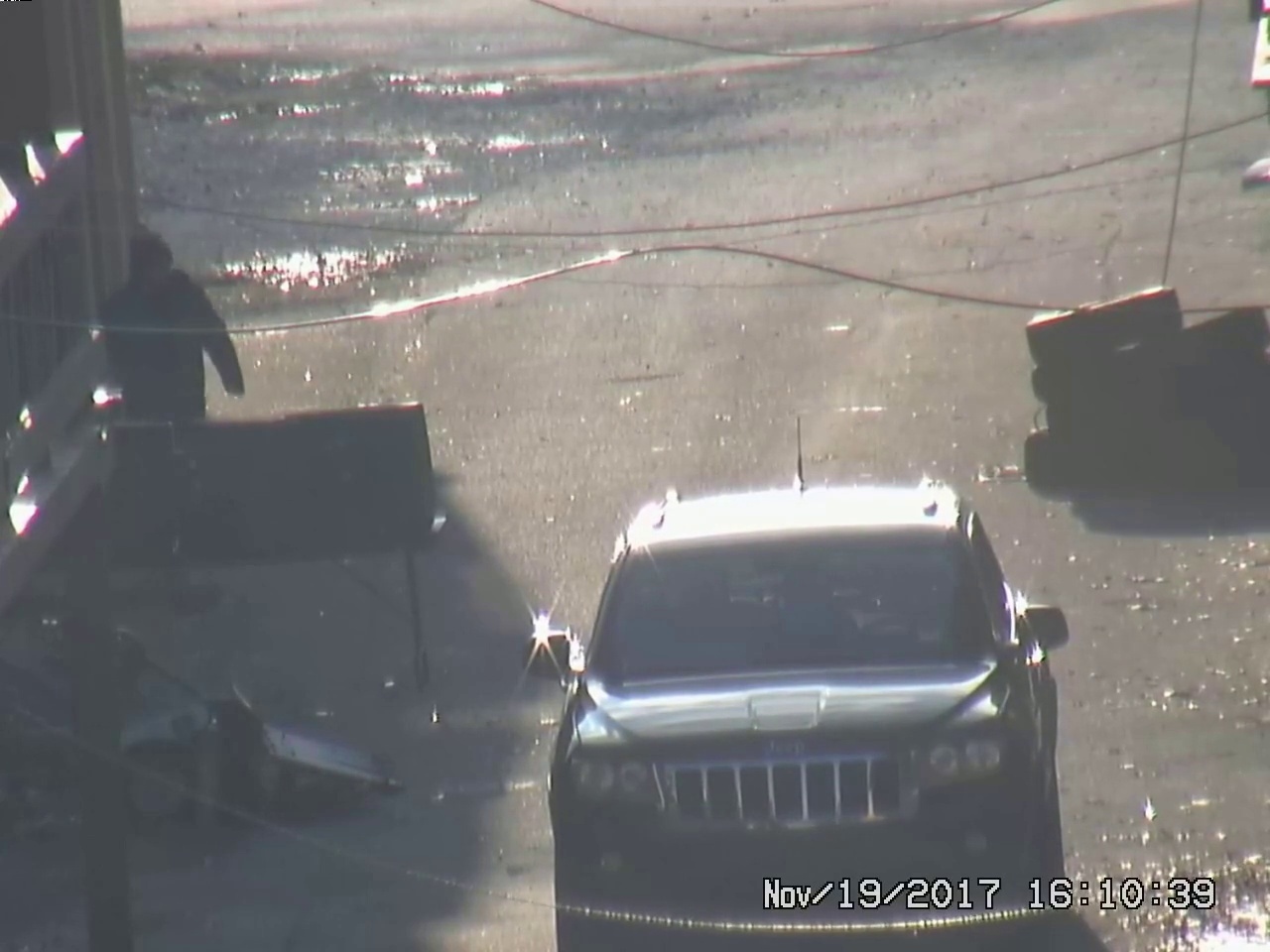}  
        &
        \includegraphics[width=0.19\textwidth]{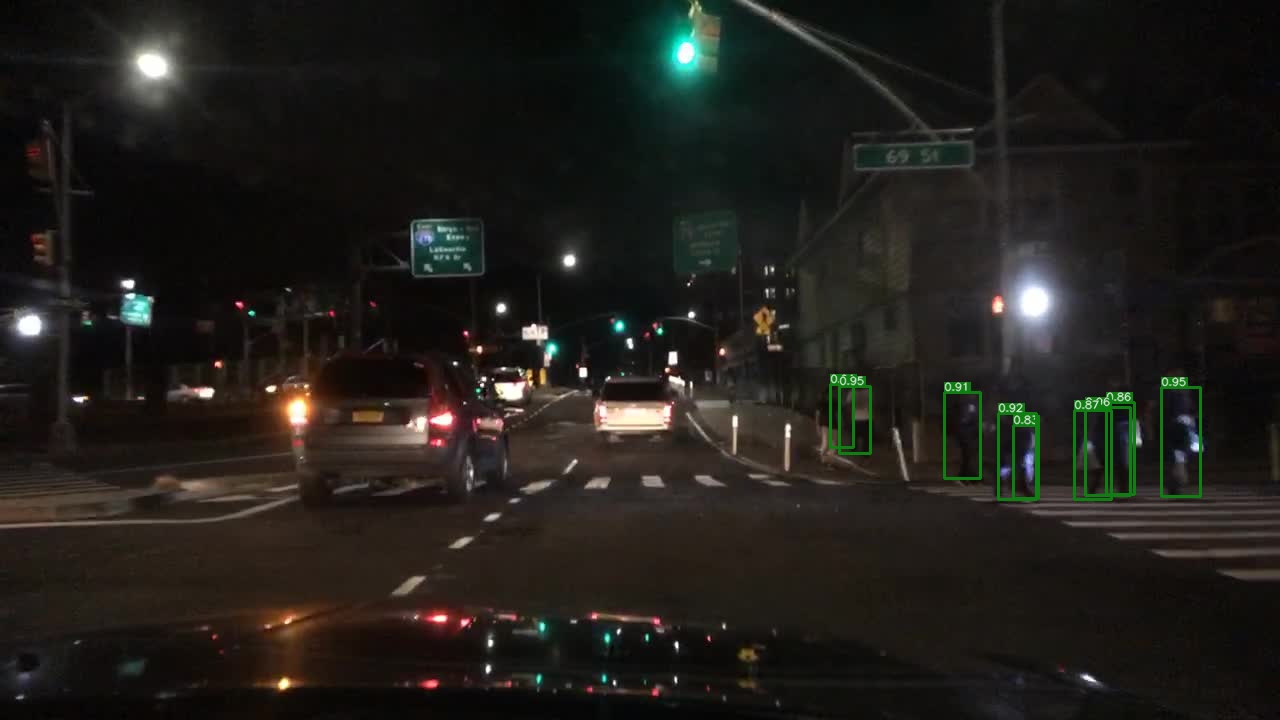} 
        &
        \includegraphics[width=0.19\textwidth]{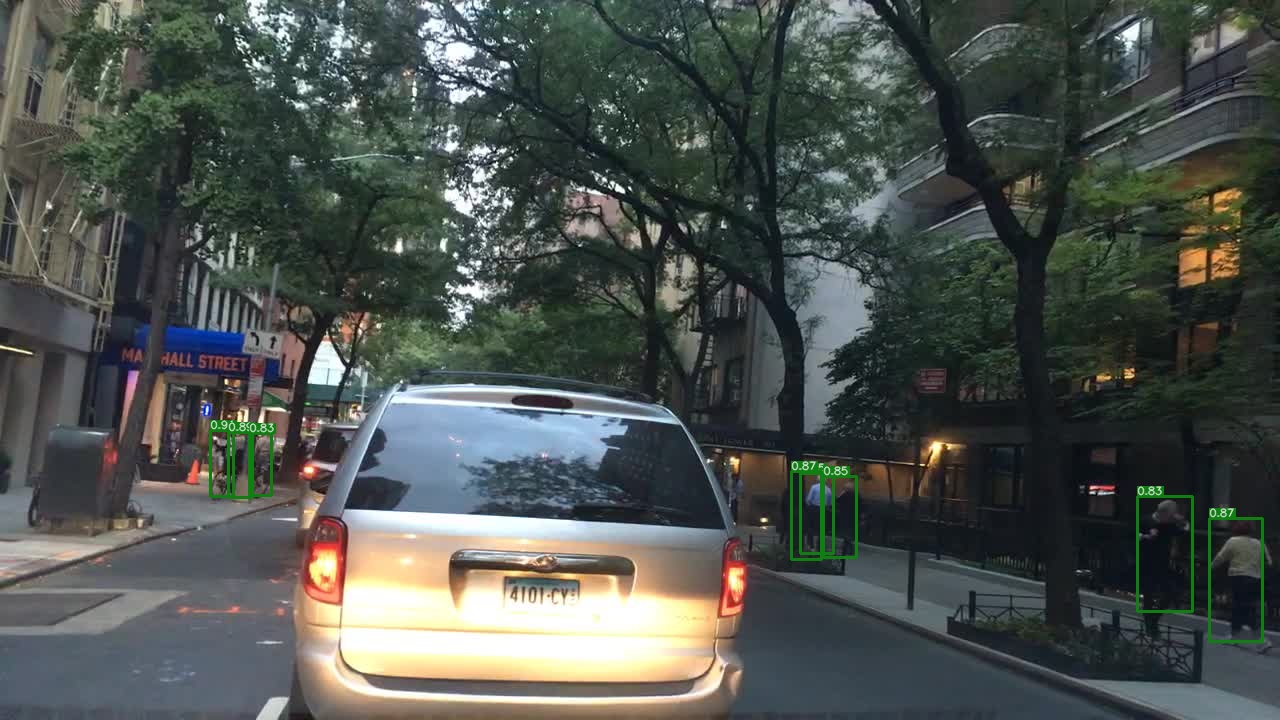} 
        &
        \includegraphics[width=0.19\textwidth]{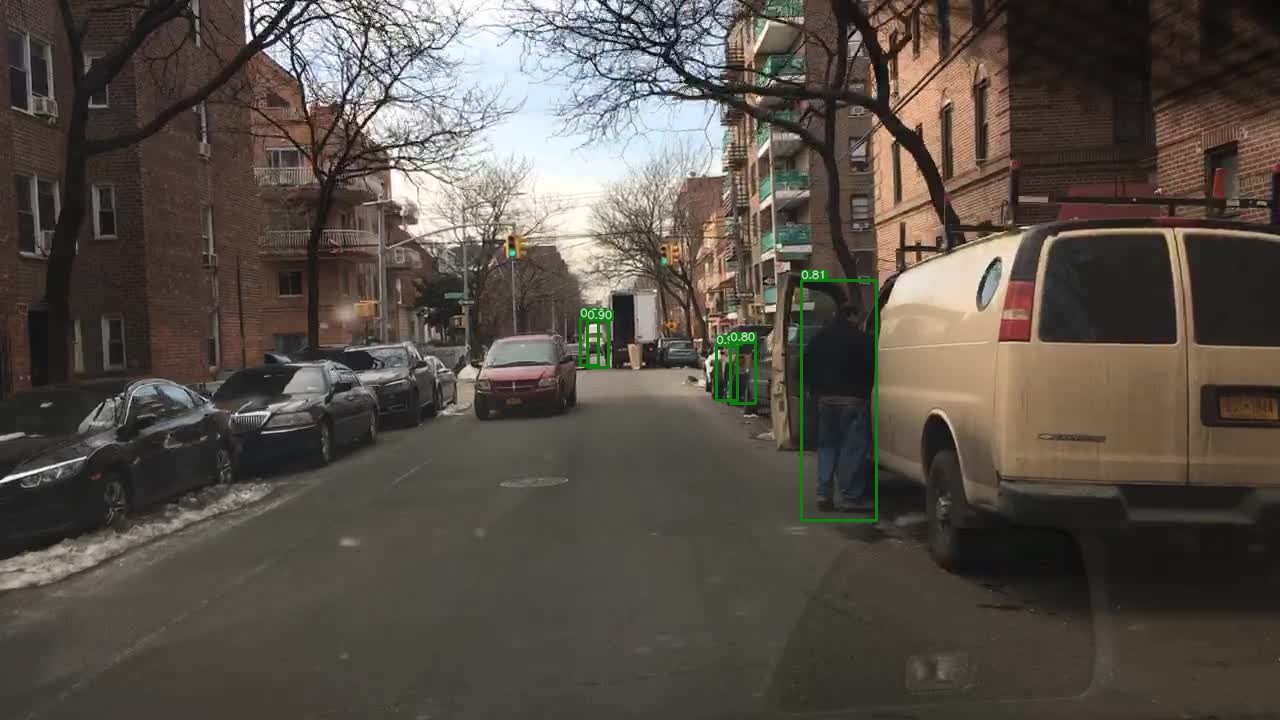} 
    \\
\end{tabular}
\vspace{-2mm}
\caption{ \textbf{Qualitative results}(best zoomed-in)\textbf{.} (a) Baseline; (b) HP~\cite{jin2018unsup}; (c) \textit{Ours}; (d) DA\cite{chen2018domain}. The domain adapted methods pick up prominent objects missed by the baseline (\textit{cols 1,3-5}). On pedestrians (\textit{cols 3-5}) the detection scores from DA is usually lower than our models', leading to lower overall performance despite correct localization. Please refer to the \href{http://vis-www.cs.umass.edu/unsupVideo/docs/self-train_cvpr2019_supp.pdf}{supplementary} for more visualizations.}
\label{fig:viz}
\vspace{-4mm}
\end{figure*}

\subsection{Face detection results}
\label{sec:face_result}
\noindent
The results on adapting from labeled WIDER Faces still-images to unlabeled CS6 surveillance video imagery are shown in Table~\ref{tab:cs6_res}.

\noindent
\textbf{Effect of pseudo-labels.}
The \textit{baseline} detector, trained on WIDER Face, gets an AP of 15.66 on CS6-Test, which underscores the domain shift between WIDER and the surveillance video domain. 
Using only the high-confidence detections ($\theta$=0.5) as training samples, \textit{CS6-}\texttt{Det}, boosts performance to 17.29 AP. Using only samples from the tracker and ignoring all pseudo-labels from the detector, \textit{CS6}-\texttt{Track}, brings down the performance to 11.73 AP. This can be partly attributed to the fact that we may miss a lot of actual faces in an image if we choose to train only on faces picked up by tracking alone. The combination of both tracking and detection results for training, \textit{CS6}-\texttt{HP}, gives slightly better performance of 17.31 AP. This is a significant boost over the model trained on WIDER-Face: $15.66 \rightarrow 17.31$.

\noindent
\textbf{Effect of soft-labels.}
Incorporating soft target labels gives a consistent gain over the default hard labels, as seen in the \texttt{Label-smooth} numbers in Table~\ref{tab:cs6_res}. 
The effect of varying the distillation weight $\lambda$ results in some fluctuation in performance -- $\text{AP}_{\lambda=0.3}$ is 19.89, $\text{AP}_{\lambda=0.5}$ is 19.56 and $\text{AP}_{\lambda=0.7}$ is 20.80. Using the completely parameter-free methods we get 19.12 from score histogram remapping (\texttt{score-remap}) and a slightly higher number, 20.65, from \texttt{HP}\textit{-cons}. Both are comparable to distillation with $\lambda=0.7$.

\noindent
\textbf{Comparison to domain discriminator.}
The domain adversarial method (DA) gives a high performance on CS6 Test with an AP of 21.02 at the image-level (\texttt{DA}-\textit{im}) and 22.18 with the instance-level adaptation included (\texttt{DA}-\textit{im-roi}). Our best numbers (20.80, 20.65) are comparable to this, given the variance over 5 rounds of training.

\begin{table}[htbp]
\renewcommand{\tabcolsep}{5pt}
\centering
\small 
\caption{\textbf{WIDER $\rightarrow$ CS6.} Average precision (AP) on  of the CS6 surveillance videos, reported as mean and standard deviation over 5 rounds of training.
}
\vspace{-3mm}
\label{tab:cs6_res}
\begin{tabular}{@{\extracolsep{5pt}}lc}
\toprule
 \textbf{Method}            & \textbf{AP} (mean $\pm$ std) \\
\midrule
Baseline: WIDER             & 15.66 $\pm$ 0.00 \\
\midrule
CS6-Det                     & 17.29 $\pm$ 0.85 \\
CS6-Track                   & 11.73 $\pm$ 0.77 \\
CS6-HP~\cite{jin2018unsup}  & 17.31 $\pm$ 0.60 \\
\midrule
CS6-Label-smooth($\lambda=0.3$)          & 19.89 $\pm$ 0.92 \\
CS6-Label-smooth($\lambda=0.5$)          & 19.56 $\pm$ 1.53 \\
CS6-Label-smooth($\lambda=0.7$)          & \textbf{20.80 $\pm$ 1.34} \\
\textit{Ours:} CS6-score-remap          & 19.12 $\pm$ 1.29 \\
\textit{Ours:} CS6-HP-\textit{cons}               & \textbf{20.65 $\pm$ 1.62}  \\
\midrule
CS6-DA-\textit{im}~\cite{chen2018domain}                   & 21.02 $\pm$ 0.96 \\
CS6-DA-\textit{im-roi}~\cite{chen2018domain}               & \textbf{22.18 $\pm$ 1.20} \\
\bottomrule
\vspace{-5mm}
\end{tabular}
\end{table}

\subsection{Pedestrian detection results}
\label{sec:ped_result}
\noindent
The results on adapting from BDD-Source images from clear, daytime videos to unconstrained settings in BDD-Target are shown in Table~\ref{tab:bdd_res}. In addition to a new task, the target domain of BDD-Pedestrians provides a more challenging situation than CS6. The target domain now consists of multiple modes of appearance -- snowy, rainy, cloudy, night-time, dusk, \etc; and various combinations thereof.

\noindent 
\textbf{Effect of pseudo-labels.}
The \textit{baseline} model gets a fairly low AP of 15.21, which is reasonable given the large domain shift from source to target.  \textit{BDD}-\texttt{Det}, which involves training with only the high-confidence detections (threshold $\theta = 0.8$), improves significantly over the baseline with an AP of 26.16. 
Using only the tracker results as pseudo-labels, \textit{BDD}-\texttt{Track}, gives similar performance (26.28). \textit{BDD}-\texttt{HP}, which combines pseudo-labels from both detection and tracking, gives the best performance among these (27.11). This is a significant boost over the baseline: 15.21 $\rightarrow$ 27.11. 

\noindent
\textbf{Effect of soft-labels.}
Using soft labels via \texttt{Label-smooth} improves results further (27.11 $\rightarrow$ 28.59), with performance fluctuating slightly with different values of the $\lambda$ hyper-parameter -- $\text{AP}_{\lambda=0.3}$ is 28.59, $\text{AP}_{\lambda=0.5}$ is 28.38 and $\text{AP}_{\lambda=0.7}$ is 28.47. Creating soft-labels via score histogram matching (\texttt{score-remap}), we get an AP of 28.02. Emphasizing tracker-only samples while constraining identical behaviour on detector training samples (\texttt{HP}-\textit{cons}) gives 28.43. Again, both these methods are comparable in performance to using \texttt{Label-smooth}, with the advantage of not having to set the $\lambda$ hyper-parameter.

\noindent
\textbf{Comparison to domain discriminator.}
Adapting to the BDD-Target domain was challenging for the domain adversarial (DA) models~\cite{chen2018domain}, most likely due to the multiple complex appearance changes, unlike the WIDER$\rightarrow$CS6 shift which has a more homogeneous target domain. The image-level adaptation (\texttt{DA}-\textit{im}) models gave 23.65 AP -- a significant improvement over the baseline AP of 15.21. We had difficulties getting the \texttt{DA}-\textit{im-roi} model to converge during training. Using the pseudo-labels from BDD-HP for class-balanced sampling of the ROIs during training had a stabilizing effect (denoted by BDD-\texttt{DA}-\textit{im-roi}*). This gives 23.69 AP. 
Overall our results from training with soft pseudo-labels are better than \cite{chen2018domain} on this dataset by $\sim$5 in terms of AP.

\begin{table}[h]
\renewcommand{\tabcolsep}{5pt}
\centering
\small
\caption{\textbf{BDD(\textit{clear,daytime}) $\rightarrow$ BDD(\textit{rest}).} Average precision (AP) on the evaluation set of the BDD pedestrian videos, reported as mean and standard deviation over 5 rounds of training.
}
\vspace{-3mm}
\label{tab:bdd_res}
\begin{tabular}{@{\extracolsep{5pt}}lc}
\toprule
 \textbf{Method}                             & \textbf{AP} (mean $\pm$ std)       \\
\midrule
Baseline: BDD(\textit{clear,daytime})        &  15.21 $\pm$ 0.00  \\
\midrule
BDD-Det                                      &  26.16 $\pm$ 0.24    \\
BDD-Track                                    &  26.28 $\pm$ 0.35   \\
BDD-HP~\cite{jin2018unsup}                   &  27.11 $\pm$ 0.54 \\
\midrule
BDD-Label-smooth($\lambda=0.3$)                              &  \textbf{28.59 $\pm$ 0.67}    \\
BDD-Label-smooth($\lambda=0.5$)                              &  28.38 $\pm$ 0.62    \\
BDD-Label-smooth($\lambda=0.7$)                              &  28.47 $\pm$ 0.41  \\
\textit{Ours:} BDD-score-remap                              &  \textit{28.02 $\pm$ 0.32}  \\
\textit{Ours:} BDD-HP-\textit{cons}                                &  \textbf{28.43 $\pm$ 0.51}  \\
\midrule
BDD-DA-\textit{im}~\cite{chen2018domain}     & 23.65 $\pm$ 0.57       \\
BDD-DA-\textit{im-roi}*                      & \textbf{23.69 $\pm$ 0.93}     \\
\bottomrule
\end{tabular}
\end{table}

\begin{figure}[h]
\centering
\small
\includegraphics[width=0.35\textwidth]{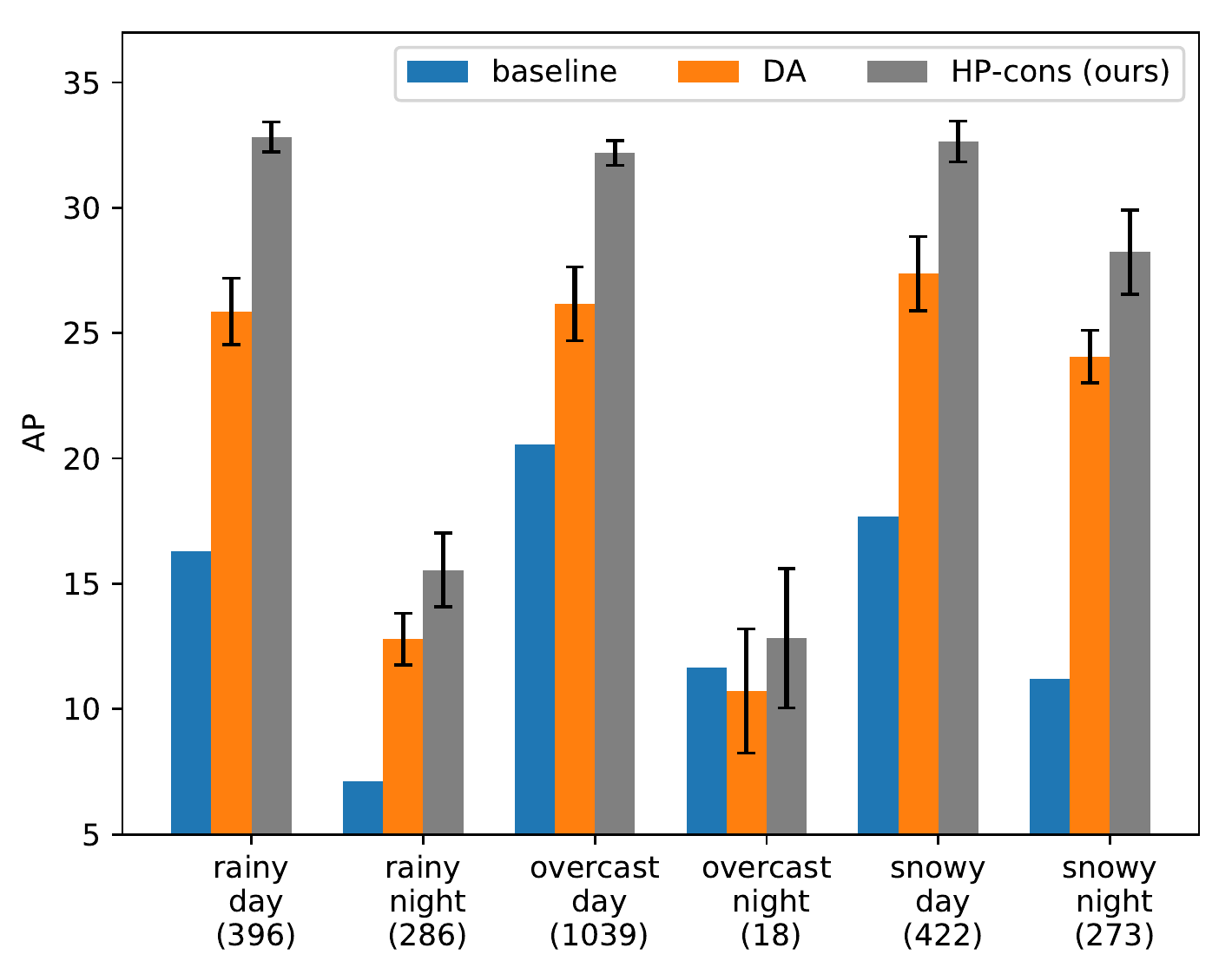} 
\vspace{-0.4cm}
\caption{\textbf{BDD(\textit{rest}) sub-domains.} Performance of the \textit{baseline} model, domain adversarial model (\textit{DA}) and our method (\textit{HP-cons}). The number of images in each sub-domain is written in parentheses below.}
\label{fig:bdd_sub}
\vspace{-3mm}
\end{figure}

\noindent
\textbf{Results on sub-domains.} 
The BDD-Target domain implicitly contains a large number of \textit{sub-domains} such as rainy, foggy, night-time, dusk, \etc. We compare the performance of three representative models -- baseline, domain adversarial (\texttt{DA}-\textit{im}) and our soft-labeling method (we pick \texttt{HP}-\textit{cons} as representative) on a set of such implicit sub-domains in BDD-Target-Test for a fine-grained performance analysis (Fig.~\ref{fig:bdd_sub}). Night-time images clearly degrade performance for all the models. Overall both domain adaptive methods improve significantly over the baseline, with \texttt{HP}-\textit{cons} consistently outperforming \texttt{DA}. It is possible that higher performance from DA can be obtained by some dataset-specific tuning of hyper-parameters on a validation set of \textit{labeled} target-domain data.

\noindent
\subsection{Automatic threshold selection} 
\label{sec:threshold}
\noindent
The hyper-parameter $\theta$ that thresholds the high-confidence detections can be set without manual inspection of the target domain. We can pick a threshold $\theta_{\mathcal{S}}$ on \textit{labeled source} data for a desired level of precision, say 0.95. Using score histogram mapping $\mathcal{S} \rightarrow \mathcal{T}$ (Sec~\ref{sec:score_map}, Fig.~\ref{fig:score_hist}), we can map $\theta_{\mathcal{S}}$ to the \textit{unlabeled target} domain as  $\theta_{\mathcal{T}}$. 
These results are shown in Table~\ref{tab:thresh}. The thresholds selected based on visual inspection of 5 videos are 0.5 for faces (17.31 AP) and 0.8 for pedestrians (27.11 AP), as described in Sec.~\ref{sec:pseudo-label}.
The performance from automatically set $\theta_{\mathcal{S} \rightarrow \mathcal{T}}$ is very close -- AP of 16.71 on CS6 and 27.11 on BDD.

\begin{table}[h]
\centering
\small
\caption{Sensitivity to detector confidence threshold for target-domain pseudo-labels, evaluated for the \textit{HP} model. The automatically selected thresholds $\theta_{\mathcal{S} \rightarrow \mathcal{T}}$ are 0.66 for CS6 and 0.81 for BDD.}
\vspace{-3mm}
    \begin{tabular}{c|cccccc}
    \toprule
    \textbf{$\theta \rightarrow$} & 0.5     & 0.6       & 0.7       & 0.8       & 0.9       & $\theta_{\mathcal{S} \rightarrow \mathcal{T}}$ \\ 
    \midrule
    \textbf{CS6-Test}                  & 17.31   & 15.91     & 14.93     &  15.63         & 11.69     & \em{16.71} \\
    \textbf{BDD-Test}                  & 27.23        & 27.68     & 27.30     &  27.11    & 25.85     & \em{27.11} \\
    \bottomrule
    \end{tabular}
    \label{tab:thresh}
\end{table}

\section{Conclusion}
\label{sec:conclusion}

\noindent
Our empirical analysis shows self-training with soft-labels to be at par with or better than the recent domain adversarial approach~\cite{chen2018domain} on two challenging tasks. Our method also avoids the extra layers and hyper-parameters of adversarial methods, which are difficult to tune for novel domains in a fully unsupervised scenario.
Our method significantly boosts the performance of pre-trained models on the target domain and gives a consistent improvement over assigning hard labels to pseudo-labeled target domain samples, the latter being prevalent in recent works~\cite{jin2018unsup,radosavovic2017data}. 
With minimal dependence on hyper-parameters, we believe our approach to be a readily applicable method for large-scale domain adaptation of object detectors.

\noindent \textbf{Acknowledgement.}
This material is based on research sponsored by the AFRL and DARPA under agreement number FA8750-18-2-0126. The U.S. Government is authorized to reproduce
and distribute reprints for Governmental purposes notwithstanding any copyright notation thereon. The views and conclusions contained herein are those of the authors and should not be interpreted as necessarily representing the official policies or endorsements, either expressed or implied, of the AFRL and DARPA or the U.S. Government. We acknowledge support from the MassTech Collaborative grant for funding the UMass GPU cluster. We thank Tsung-Yu Lin and Subhransu Maji for helpful discussions.




\end{document}